\documentclass[10pt,twocolumn,letterpaper]{article}

\usepackage{cvpr}
\usepackage{times}
\usepackage{epsfig}
\usepackage{graphicx}
\usepackage{amsmath}
\usepackage{amssymb}
\usepackage[numbers]{natbib}

\usepackage{algorithmic}
\usepackage{algorithm}
\usepackage{tikz}
\usepackage{float}
\usepackage{caption}
\usepackage{microtype}
\usepackage{bm}

\usepackage[T1]{fontenc}
\usepackage[utf8]{inputenc}

\usepackage{mathtools}

\usepackage[pagebackref=true,breaklinks=true,letterpaper=true,colorlinks,bookmarks=false]{hyperref}

\cvprfinalcopy 

\def\cvprPaperID{1848} 

\ifcvprfinal\fi
\begin{document}

\title{Transflow Learning: Repurposing Flow Models Without Retraining}

\author{Andrew Gambardella\\
University of Oxford\\
{\tt\small gambs@robots.ox.ac.uk}
\and
Atılım Güneş Baydin \\
University of Oxford\\
{\tt\small gunes@robots.ox.ac.uk}
\and
Philip H. S. Torr \\
University of Oxford \\
{\tt\small phst@robots.ox.ac.uk}
}

\maketitle

\begin{abstract}
    It is well known that deep generative models have a rich latent space, and that it is possible to smoothly manipulate their outputs by traversing this latent space. Recently, architectures have emerged that allow for more complex manipulations, such as making an image look as though it were from a different class, or painted in a certain style. These methods typically require large amounts of training in order to learn a single class of manipulations. We present Transflow Learning, a method for transforming a pre-trained generative model so that its outputs more closely resemble data that we provide afterwards. In contrast to previous methods, Transflow Learning does not require any training at all, and instead warps the probability distribution from which we sample latent vectors using Bayesian inference. Transflow Learning can be used to solve a wide variety of tasks, such as neural style transfer and few-shot classification.
\end{abstract}

\begin{figure}[h]
\begin{center}
\includegraphics[width=.2\textwidth]{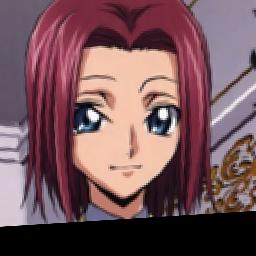}
\includegraphics[width=.2\textwidth]{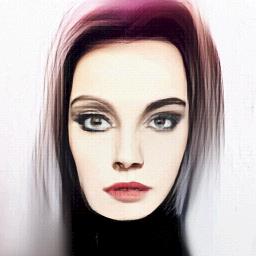}
\\
\includegraphics[width=.2\textwidth]{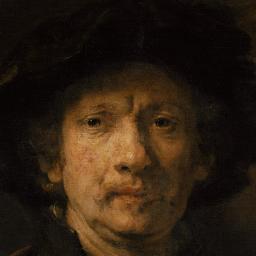}
\includegraphics[width=.2\textwidth]{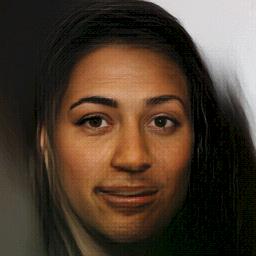}
   \caption{Transflow Learning allows us to warp the latent distribution of any trained invertible generative model, so that we can instead sample data similar to that we provide post-hoc. This works by treating the latent distribution as the prior in a Bayesian posterior inference setting where we condition on the provided data. In this example, given only 18--24 instances of images in the art domains on the left, a flow model trained on CelebA \citep{Liu2014} is able to generate human faces with matching attributes, even though these attributes are not contained in the CelebA dataset.}
\end{center}
\end{figure}

\section{Introduction}

One of the greatest challenges in modern machine learning is few-shot learning \citep{koch2015siamese,lake2011one}. Whereas a human is capable of learning a task such as handwritten digit recognition after only having seen a few samples of each digit, even simple machine learning classifiers require training multiple epochs over a relatively large dataset. When it comes to more complicated tasks, machine learning algorithms become even more data-hungry---whereas humans can learn to play Atari games in a matter of minutes, even the most sample-efficient of reinforcement learning algorithms take hundreds of hours of gameplay \citep{Tsividis2017}.

What advantages do humans have over machines that allows us to consistently beat them with regards to sample efficiency? One might argue that humans are constantly utilizing their experiences in other domains in order to draw parallels between tasks. Even when it comes to very disparate sets of tasks, such as ``natural language understanding" and ``video game playing," studies have shown that it is possible to transfer knowledge between these domains for major sample efficiency gains in both machine learning algorithms and human learning \citep{Tsividis2017, Branavan2014}. The idea of taking knowledge from one domain or task, and using that knowledge in another domain, is known as ``transfer learning'' \citep{pan2009survey}.

When attempting to transfer knowledge about one task to another, two broad questions must be asked: how is the prior knowledge from other tasks stored, and how can it be used? We propose an answer to these questions in the context of generative models.

Modern generative models, such as Generative Adversarial Networks (GANs) \citep{Goodfellow2014}, normalizing flow models \citep{Rezende2015}, and autoregressive models such as Conditional PixelCNN \citep{Oord2016} differ in their learning mechanisms, but all share a common thread: they learn a function $f_{\bm{\theta}}:\mathbb{R}^d\to\mathbb{R}^d$ which transforms samples of a latent random variable $\mathbf{z}\sim q(\mathbf{z})$, where $\mathbf{z}\in\mathbb{R}^d$ and $q(\mathbf{z})$ is a known distribution, to a data point (\eg , an image) $\mathbf{x}\in\mathbb{R}^d$. The latent distribution $q(\mathbf{z})$ is usually a simple distribution such as the multivariate Gaussian, $\mathcal{N}(\vec{0},I)$. When learning generative models we keep $q(\mathbf{z})$ fixed, and concentrate all of our efforts on optimizing the parameters $\bm{\theta}$.

In the context of above discussion on transfer learning, however, we take a different view. We wish to transfer knowledge from a trained generative model $f$, to some new task. To be concrete, assume we have a generative model that will output an arbitrary celebrity face when given a latent vector $\mathbf{z} \sim \mathcal{N}(\vec{0},I)$ as input. Can we use the same generative model to only output celebrities with red hair? Can we output a celebrity which looks like an anime character? Can we even classify handwritten digits?

We find that all of the above and more is possible, with the condition that our generative model $f$ is \emph{invertible}. That is, we require an inverse function $\mathbf{z}=f^{-1}(\mathbf{x})$ which takes a data point (\eg , an image) as input, and outputs the corresponding latent vector. Normalizing flow models \citep{Rezende2015} are the most natural class of such generative models, as they are by nature invertible, unlike other architectures such as GANs which can be inverted only under specific circumstances \citep{Donahue2016}. 

Our method works by taking the flow model and its parameters as fixed, and instead warping the latent distribution so that we can control the latent vectors that we sample. Specifically, we treat the latent distribution $q(\mathbf{z})$ as a prior distribution in a Bayesian inference setting where we update it to a posterior $q(\mathbf{z}\vert \bm{\zeta}_{1:m})$ conditioned on some observed data samples $\mathbf{x}_{1:m}$ mapped to corresponding latent vectors using $\bm{\zeta}_i=f^{-1}(\mathbf{x}_i), i=1,\dots,m$. We call our method Transflow Learning, as it uses flow models to perform tasks for which they were not originally trained.

In Section \ref{sec:preliminaries} we cover some essential concepts, followed by Section \ref{sec:algorithm} where we describe the mechanism by which we warp $q(\mathbf{z})$: Bayesian inference in the latent space. Section \ref{sec:related_work} covers related work. In Section \ref{sec:experiments}, we provide example use cases in which knowledge can be transferred, specifically modeling distributions other than the training data, and solving downstream tasks. Section \ref{sec:future_work} discusses future work followed by conclusions in Section \ref{sec:conclusions}.


\section{Background}
\label{sec:preliminaries}

\subsection{Normalizing Flow Models}

A normalizing flow is a series of learned invertible transformations which can transform one probability distribution into another. If random variable $\mathbf{z}_0$ with associated probability density $q_0(\mathbf{z}_0)$ is put through a series of invertible transformations $\{f_1, ..., f_K\}$ so that
\begin{gather}
\mathbf{z}_K = f_K \circ ... \circ f_1(\mathbf{z}_0) = f(\mathbf{z}_0)
\end{gather}
then we have
\begin{gather}
q_K(\mathbf{z}_K) = q_0(\mathbf{z}_0) \prod_{k=1}^K \left|{\det \frac{\partial f_k}{\partial \mathbf{z}_{k-1}}}\right|^{-1}\; .
\end{gather}

Each $f_k$ contains learnable parameters, which are typically learned by maximum likelihood. The flow is ``normalizing" because each $q_k(\mathbf{z}_k)$ defines a valid probability distribution \citep{Rezende2015}.

In this paper we designate $\mathbf{z}=\mathbf{z}_0$ as the latent vector and $\mathbf{x}=\mathbf{z}_K$ as the data point produced by the transformation $\mathbf{x}=f(\mathbf{z})$. We require two properties of flow models: that $f$ is invertible (which is true as it is the composition of invertible functions), and that vectors around $f^{-1}(\mathbf{x})$ correspond to data close to $\mathbf{x}$, even for $\mathbf{x}$ that the flow model had not seen during training (which holds empirically as long as $\mathbf{x}$ is not extremely unnatural).

\subsection{Posterior Inference}

Bayesian inference is a powerful tool for reasoning about probability distributions \citep{gelman2013bayesian}. Core to the idea of Bayesian inference are the concepts of the \emph{prior}, which is a probability distribution $p(z)$ that we assume exists before having seen any data, and the \emph{posterior}, which is a probability distribution $p(z\vert x)$ that we obtain after having observed some data (or evidence) $x$ with a \emph{likelihood} $p(x\vert z)$. These are related with the relationship $p(z\vert x)=p(x\vert z)\,p(z)/p(x)$. The likelihood is essentially a weighting that tells us to what degree the prior must be moved after having observed some evidence.



 Also important is the concept of a \emph{conjugate prior} \citep{schlaifer1961applied}, which means that with certain choices of prior and likelihood distributions, we can ensure that we have a closed-form analytical expression for the posterior distribution. In particular, we will need to use the fact that if we have a prior which is a multivariate Gaussian and a likelihood function which is a multivariate Gaussian with a known covariance matrix, then the posterior is also guaranteed to be a multivariate Gaussian.

Using knowledge of conjugate distributions is particularly attractive, as it will allow us to solve for the posterior parameters analytically. Without a certain specification of likelihood function, we would need to resort to sampling-based methods such as stochastic variational inference \citep{Hoffman2012} in order to obtain an approximation to the posterior.


\section{Algorithm}
\label{sec:algorithm}

\begin{algorithm}[tb]
    \caption{Transflow Learning}
    \label{alg:tflearning}
    \begin{algorithmic}[1]
       \STATE {\bfseries Input:} Trained flow model $\mathbf{x}=f(\mathbf{z})$ with inverse $\mathbf{z}=f^{-1}(\mathbf{x})$, where $\mathbf{x}$ are data and $\mathbf{z}$ are latents
       \STATE For each set of evidence data $\mathbf{x}_{1:m}$, find corresponding latents $\{\bm{\zeta}_1, ..., \bm{\zeta}_m\} = \{f^{-1}(\mathbf{x}_1), ..., f^{-1}(\mathbf{x}_m)\}$
       \STATE Construct posterior $q(\mathbf{z}|\bm{\zeta}_{1:m})=\mathcal{N}(\mathbf{\mu}_p, \mathbf{\Sigma}_p)$ by computing $\bm{\mu}_p$ and $\mathbf{\Sigma}_p$ analytically
       \STATE Obtain samples from the data posterior $p(\mathbf{x}|\mathbf{x}_{1:m})$ by computing $\mathbf{x}=f(\mathbf{z})$ where $\mathbf{z} \sim q(\mathbf{z}|\bm{\zeta}_{1:m})$
    \end{algorithmic}
    \label{algo}
\end{algorithm}

\begin{figure}
\begin{center}
\begin{minipage}{.23\textwidth}
\includegraphics[width=\textwidth]{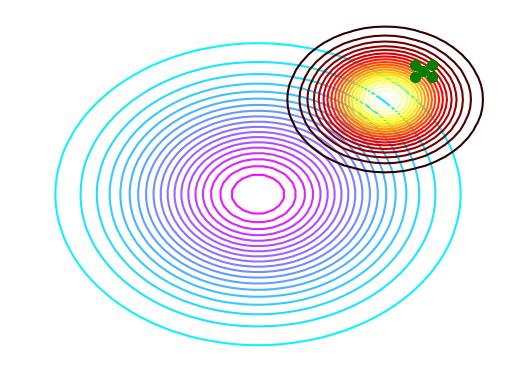}
\caption*{$\lambda = 0.3m$}
\end{minipage}
\begin{minipage}{.23\textwidth}
\includegraphics[width=\textwidth]{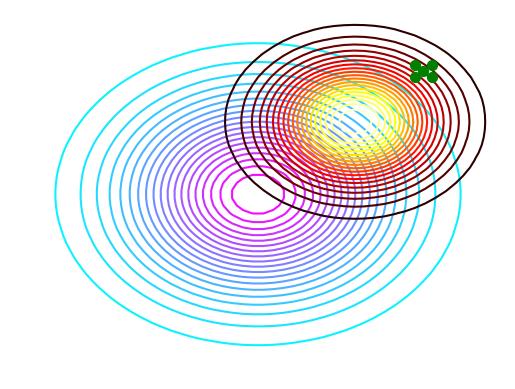}
\caption*{$\lambda = 0.7m$}
\end{minipage}
\end{center}
\caption{Transflow Learning finds a posterior (top right) in between the prior (bottom left) and the evidence (cross mark). We see that as $\lambda$ becomes larger, the mean of the posterior becomes closer to the mean of the prior, and the covariance of the posterior becomes larger, but in both cases the evidence can be sampled with relatively high probability.}
\end{figure}

Our algorithm is detailed in Algorithm \ref{alg:tflearning}. 
The key insight is to treat the underlying latent distribution $q(\mathbf{z})$ of a flow model as a prior, where usually $q(\mathbf{z})=\mathcal{N}(\vec{0},I)$. We are then interested in obtaining a posterior over the latent space of the flow model $q(\mathbf{z}\vert\bm{\zeta}_{1:m})$, conditioned on $\bm{\zeta}_i$, which are some data observations $\mathbf{x}_i$ mapped to the latent space using $\bm{\zeta}_i=f^{-1}(\mathbf{x}_i),i=1,\dots,m$. 

In other words, we provide evidence in the form of new latent vectors and, conditioned on this evidence, we find a posterior distribution over the flow model's latent variables. This effectively gives us a new generative model from which we can sample data resembling the evidence. In order to accomplish this, we require our generative model $f$ to be invertible.

\subsection{Computing the Posterior over Latent Vectors}
\label{sec:posterior_over_latent_vectors}
As most implementations of normalizing flow models use a multivariate Gaussian to model $q(\mathbf{z})$, with an appropriate choice of likelihood function we can compute the posterior analytically. Assume we are given a trained flow model, and that during training the model was shown latent vectors from $\mathbf{z} \sim \mathcal{N}(\vec{0},I)$. If we use a multivariate Gaussian likelihood function with covariance matrix $\Sigma$, then the posterior over the latent vectors is also a multivariate Gaussian, so that $q(\mathbf{z}\vert\bm{\zeta}_{1:m})=\mathcal{N}(\bm{\mu}_p,\bm{\Sigma}_p)$, and its parameters are given by the formulae:
\begin{gather}
\bm{\mu}_p = (\bm{\Sigma}^{-1}_0 + m \bm{\Sigma}^{-1})^{-1} (\bm{\Sigma}^{-1}_0 \bm{\mu}_0 + m \bm{\Sigma}^{-1}\bar{\bm{\zeta}})\\
\bm{\Sigma}_p = (\bm{\Sigma}_0^{-1} + m \bm{\Sigma}^{-1})^{-1}
\end{gather}

\noindent
where $m$ is the number of observed data points, $\bar{\bm{\zeta}}$ is the mean of observed latent vectors $\bm{\zeta}_{1:m}$, $\bm{\mu}_0$ is the prior mean and $\bm{\Sigma}_0$ is the prior covariance matrix. As we know that $\bm{\mu}_0$ is equal to $\vec{0}$ and $\bm{\Sigma}_0$ is the identity matrix, we can further simplify these formulae:
\begin{gather}
\bm{\mu}_p = (I + m \bm{\Sigma}^{-1})^{-1} (m \bm{\Sigma}^{-1}\bar{\bm{\zeta}})\\
\bm{\Sigma}_p = (I + m \bm{\Sigma}^{-1})^{-1}
\end{gather}

\noindent
The choice of $\bm{\Sigma}$ here serves as a hyperparameter. One natural choice would be to use a scalar matrix, which implies that we would like to keep the latent dimensions uncorrelated and weighted identically. With likelihood covariance $\bm{\Sigma}$ being set to $\lambda I$, with $\lambda$ a scaling hyperparameter, the posterior parameters become simple to compute:
\begin{gather}
\bm{\mu}_p = (I + \frac{m}{\lambda} I)^{-1} (\frac{m}{\lambda} I\bar{\bm{\zeta}}) = \frac{\frac{m}{\lambda}\bar{\bm{\zeta}}}{\frac{m}{\lambda} + 1} \\
\bm{\Sigma}_p = (I + \frac{m}{\lambda} I)^{-1} = \frac{1}{\frac{m}{\lambda} + 1} I
\end{gather}

There are three special cases that we can examine:
\begin{enumerate}
\item If we let $\lambda = c$, where $c$ is a small constant relative to $m$, then the posterior mean will be close to the sample mean, and the posterior covariance will be very small.
\item If we let $\lambda = m$, then the posterior mean will be locked onto $0.5 \bar{\bm{\zeta}}$, and the posterior covariance will remain constant at $0.5 I$, regardless the value of $m$.
\item If we let $\lambda$ become arbitrarily large, then the posterior mean will be close to $\vec{0}$ and the posterior covariance will be close to $I$, \ie , all conditioning is completely ignored and we get back the original flow model.
\end{enumerate}

In other words, low values of $\lambda$ relative to $m$ will give a posterior that is close to the sample mean, and with very low covariance, whereas high values of $\lambda$ will become more and more like the original flow model.

\subsection{Computing the Posterior Predictive Distribution}
\label{sec:post_pred}

The posterior predictive distribution evaluates the probability of a possible unobserved value $\bm{\zeta}$ of latent vectors conditioned on the observed values $\bm{\zeta}_{1:m}$. It is obtained by marginalizing the distribution of $\bm{\zeta}$ over the posterior $q(\mathbf{z}\vert\bm{\zeta}_{1:m})$:
\begin{gather}
q(\bm{\zeta}\vert\bm{\zeta}_{1:m})=\int p(\bm{\zeta}\vert\mathbf{z},\bm{\zeta}_{1:m})\,q(\mathbf{z}\vert\bm{\zeta}_{1:m})\,\mathrm{d}z
\end{gather}

In the setting of multivariate Gaussian conjugate priors that we described in the previous section, its form is known and easily calculated:
\begin{gather}
\bm{\mu}_{pp} = \bm{\mu}_p = \frac{\frac{m}{\lambda}\bar{\bm{\zeta}}}{\frac{m}{\lambda} + 1} \\
\bm{\Sigma}_{pp} = \bm{\Sigma}_p + \bm{\Sigma} = \frac{1}{\frac{m}{\lambda} + 1} I + \lambda I
\end{gather}

Its mean is exactly the same as that of the posterior, and its covariance is also identical, save for the extra $\lambda I$ term, which accounts for uncertainty in the parameters. We can use the posterior predictive distribution for a wide variety of tasks unrelated to sampling, and in Section \ref{sec:mnist} we will show how to do MNIST \citep{LeCunYann1998} classification without training.

\subsection{Properly Setting $\lambda$}

The hyperparameter $\lambda$ determines the variance of the likelihood that is used in the conditioning on observed data points. This is similar to the use of approximate Bayesian computation \citep{marjoram2003markov,wilkinson2013approximate} likelihoods in Bayesian inverse graphics \citep{mansinghka2013approximate}, where the variance of the likelihood plays the role of a ``tolerance'' in judging how closely an image generated by the generative model matches an observed image. High tolerances admit generation of images that do not closely match the observation, whereas low tolerances push inference towards closely mimicking the observation while reducing the sample efficiency in complex image settings.

From the perspective of the analytical posteriors introduced in Section \ref{sec:posterior_over_latent_vectors}, while setting $\lambda$ to a high value may seem like a mistake due to the behaviour of the posterior as $\lambda$ grows larger, we argue that low values of $\lambda$ are even more dangerous. As flow models learn invertible maps, the dimensionality of the latent vectors must be equal to that of the output. For example, if we wish to output full-colour, 256 by 256 images, then the dimensionality of the latent space is $3\times 256\times 256 = 196,608$. In contrast, the dimensionality of the latent space for a typical GAN \citep{Goodfellow2014} or VAE \citep{Kingma2013}, which do not have this restriction, is around 100.

The high dimensionality of flow model latent vectors implies that vectors which should be ``close" in that they share similar features in image space will be very far in the $L_2$ sense. This has implications for the sample mean of latent vectors, $\bar{\bm{\zeta}}$, which will have a smaller $L_2$ norm as more observed data points are averaged, due to the curse of dimensionality spreading supposedly ``similar" vectors in different directions relative to the origin. As vectors with smaller norm are closer to the mean of Gaussian on which the flow model was trained, $\vec{0}$, this has the effect that conditioning on many data points will give a posterior mean which is very ``generic," as it has unreasonably high probability under the original model (\ie , much higher probability than a vector randomly sampled from $\mathcal{N}(\vec{0},I)$). While this is not bad in and of itself (after all, the mean of the original distribution, $\vec{0}$, is the most generic image possible), this issue is further compounded by the covariance of the posterior shrinking as more data points are added, making it so that if $\lambda$ is set too low, every sample from the distribution is extremely generic. Even though the mean of the posterior distribution has even smaller norm with larger values of $\lambda$, the larger covariance makes up for it.

In practice we find that a large range of settings of $\lambda$ work, depending on the nature of the conditioning, but in general values which are in the range $[0.3m, 0.7m]$ are preferable. We explore the consequences of different choices of $\lambda$ in Section \ref{sec:experiments}.


\section{Related Work}
\label{sec:related_work}

\begin{figure*}
\begin{center}
\begin{tikzpicture}

\node[inner sep=0pt] (kallen_real) at (-6,0)
    {\includegraphics[width=.2\textwidth]{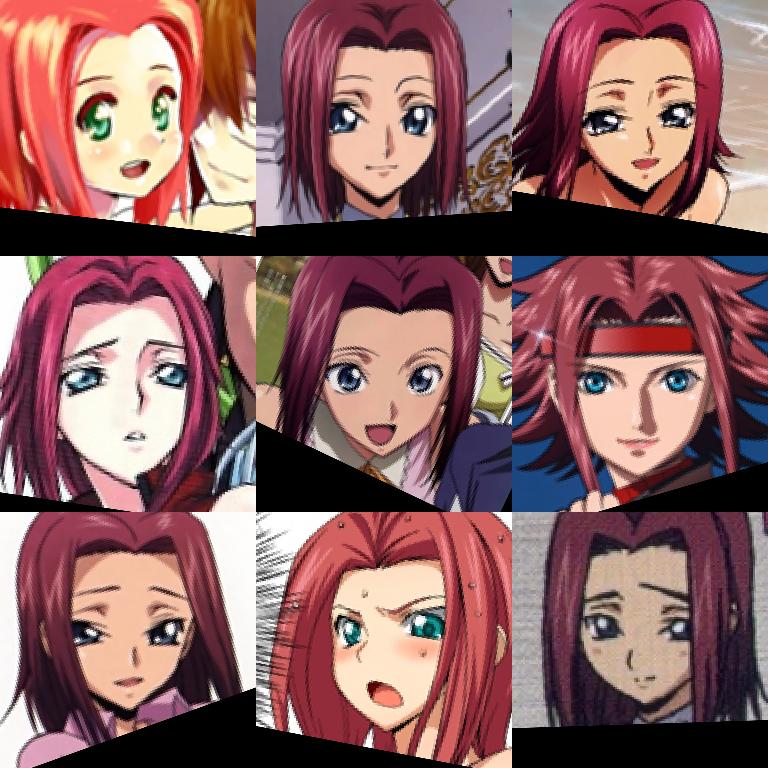}};
\node[inner sep=0pt][label={\small (1)}] (kallen_projection) at (1,0)
    {\includegraphics[width=.2\textwidth]{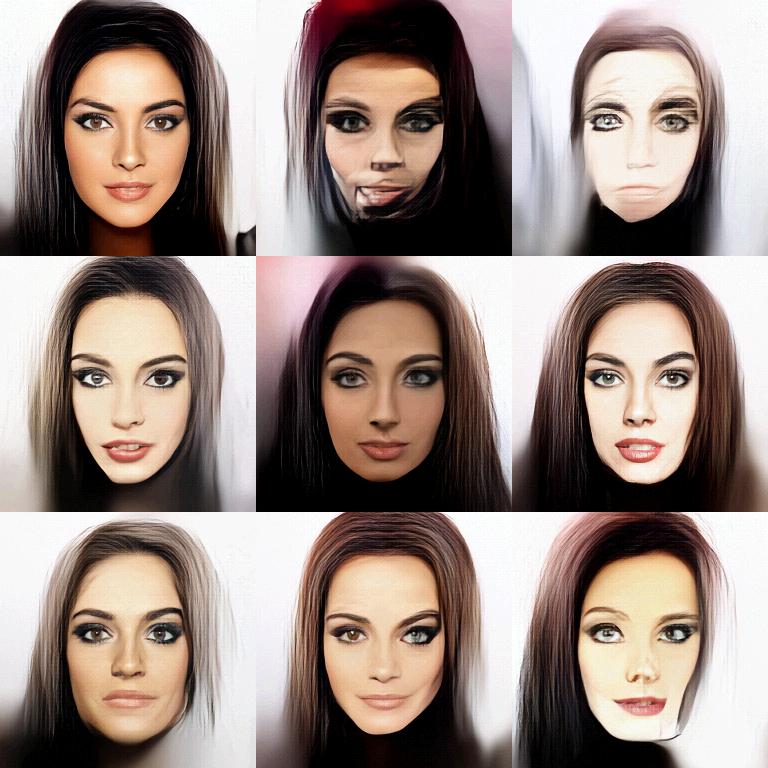}};
\node[inner sep=0pt][label={\small (2)}] (kallen_midpoint) at (4.5,0)
    {\includegraphics[width=.2\textwidth]{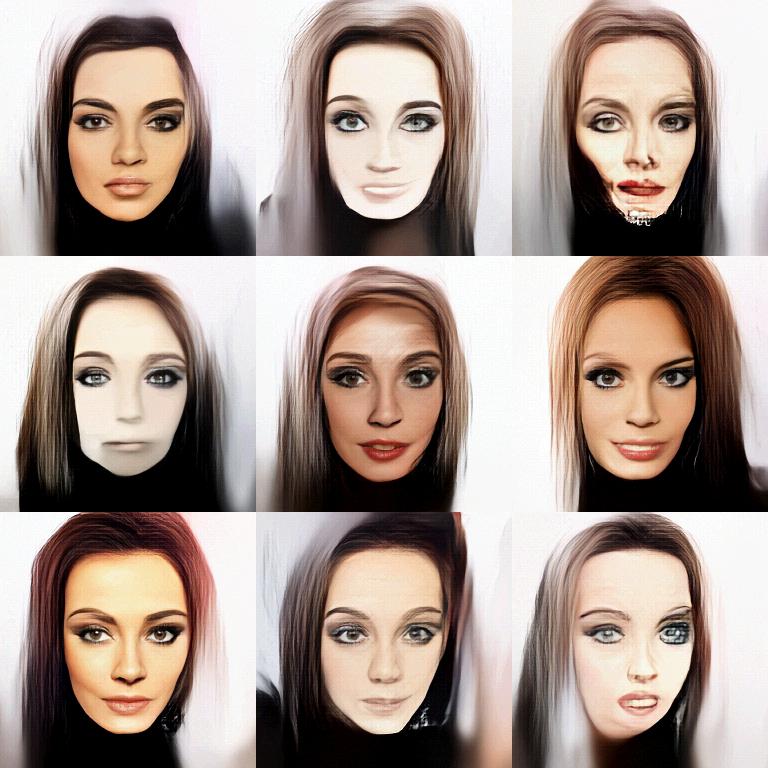}};
\node[inner sep=0pt][label={\small (3)}] (interpolation_midpoint_1) at (8,0)
    {\includegraphics[width=.2\textwidth]{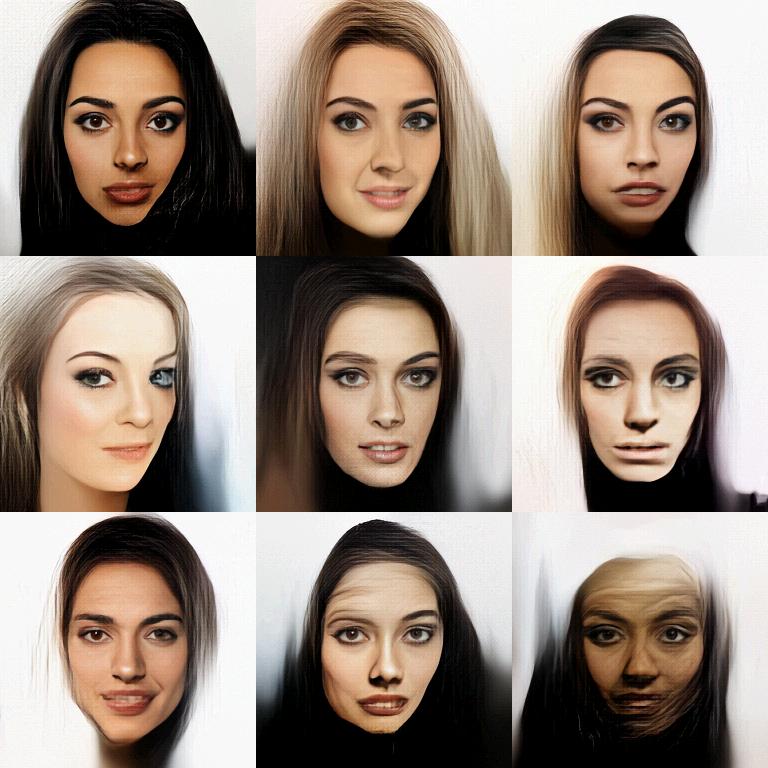}};
\node[inner sep=0pt][label=below:{\small (4)}] (interpolation_midpoint_2) at (8,-4)
    {\includegraphics[width=.2\textwidth]{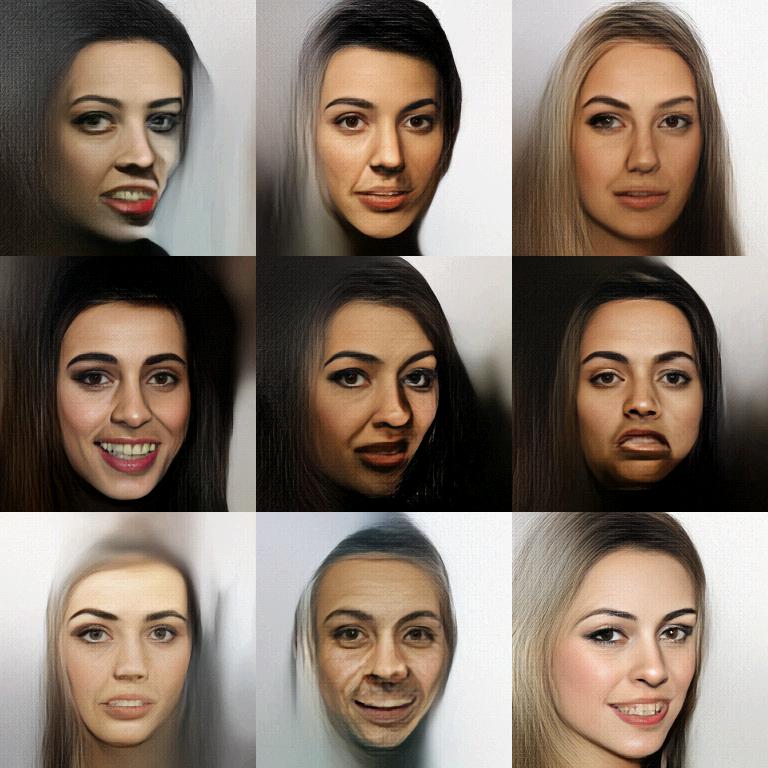}};
\node[inner sep=0pt][label=below:{\small (5)}] (rembrandt_midpoint) at (4.5,-4)
    {\includegraphics[width=.2\textwidth]{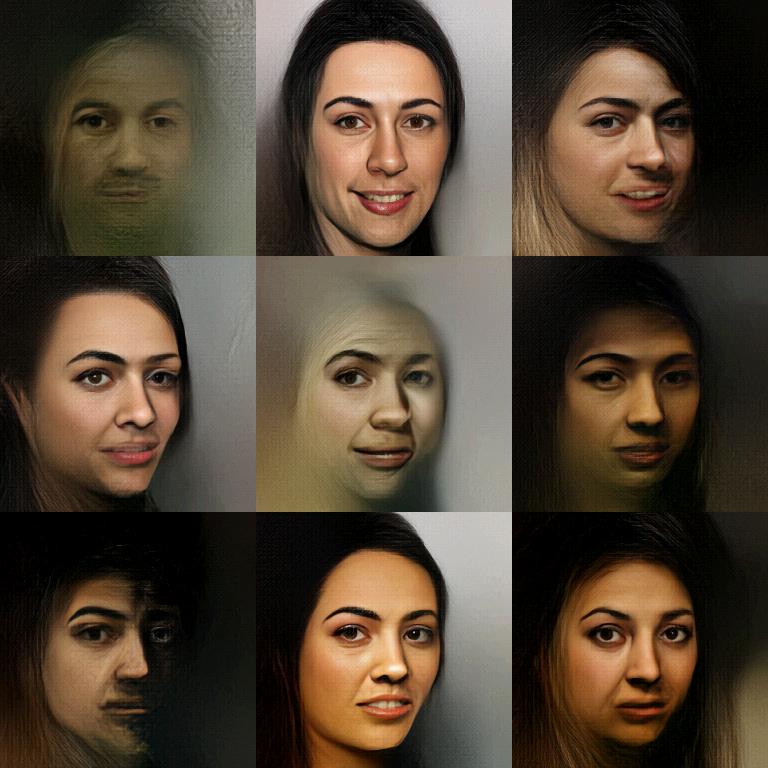}};
\node[inner sep=0pt][label=below:{\small (6)}] (rembrandt_projection) at (1,-4)
    {\includegraphics[width=.2\textwidth]{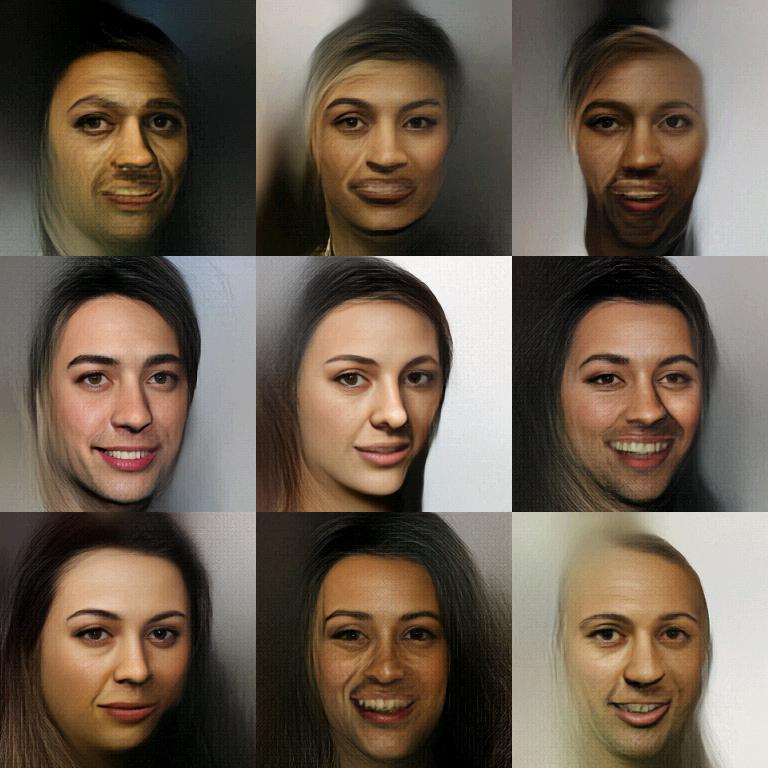}};
\node[inner sep=0pt] (rembrandt_real) at (-6,-4)
    {\includegraphics[width=.2\textwidth]{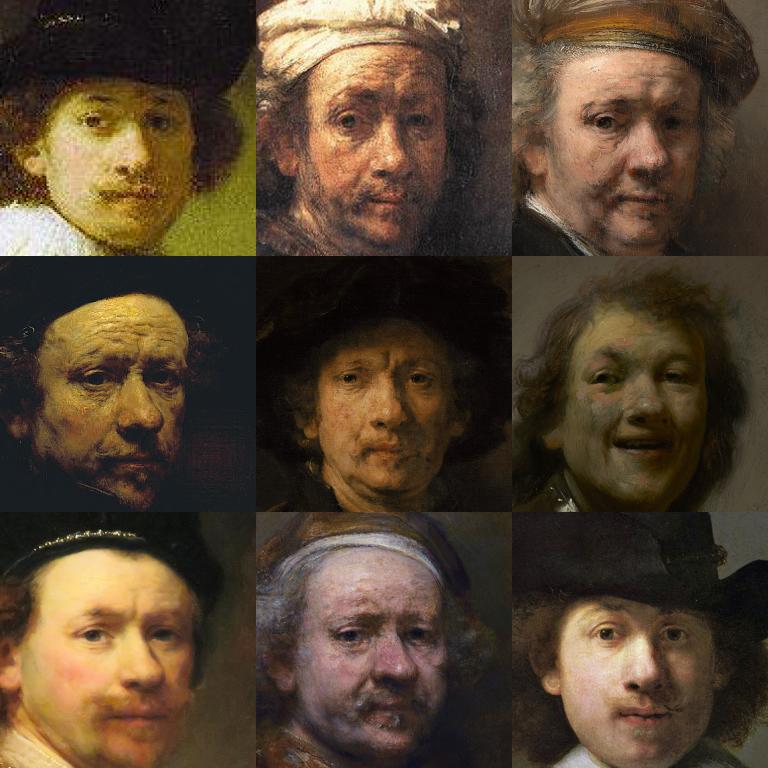}};

\draw[->,thick] (kallen_real.east) -- (kallen_projection.west)
    node[midway,fill=white] {$\mathcal{N}(\frac{\frac{m_a}{\lambda}\bar{\bm{\zeta}}_a}{\frac{m_a}{\lambda} + 1} ,\frac{1}{\frac{m_a}{\lambda} + 1} I)$};
\draw[->,thick] (rembrandt_real.east) -- (rembrandt_projection.west)
    node[midway,fill=white] {$\mathcal{N}(\frac{\frac{m_b}{\lambda}\bar{\bm{\zeta}}_b}{\frac{m_b}{\lambda} + 1} ,\frac{1}{\frac{m_b}{\lambda} + 1} I)$};

\end{tikzpicture}
\end{center}
  \caption{Interpolation between two sets of images far outside the training distribution, by first projecting onto the manifold of human faces, and then interpolating the parameters of the posterior distributions. Note that as the distribution gets closer to that of Rembrandt's self-portraits, the colours in the image get darker, men are sampled much more frequently, the hair is often gone from the samples (as Rembrandt often wore a hat which blended in with the background), and the sampled faces are more tilted towards the right. (View in numerical or reverse numerical order)}
\label{fig:interpolation}
\end{figure*}

\begin{figure*}
\begin{center}
\includegraphics[width=.09\textwidth]{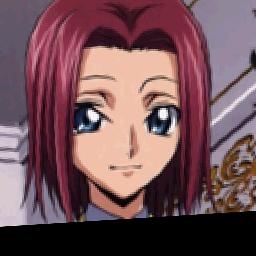}
\includegraphics[width=.09\textwidth]{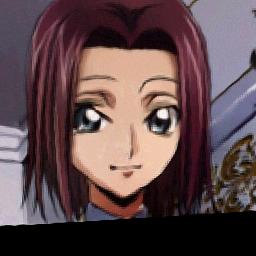}
\includegraphics[width=.09\textwidth]{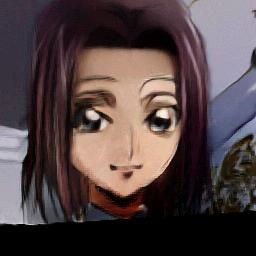}
\includegraphics[width=.09\textwidth]{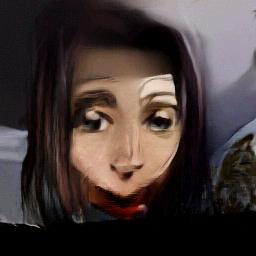}
\includegraphics[width=.09\textwidth]{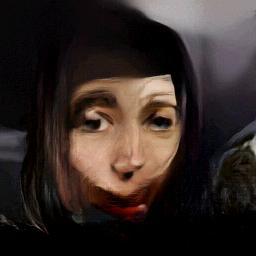}
\includegraphics[width=.09\textwidth]{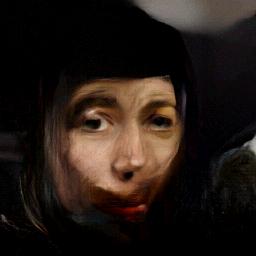}
\includegraphics[width=.09\textwidth]{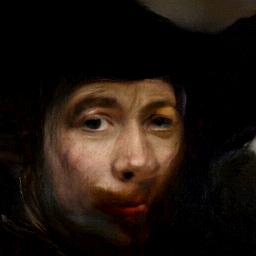}
\includegraphics[width=.09\textwidth]{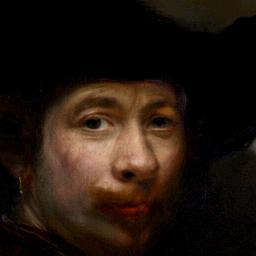}
\includegraphics[width=.09\textwidth]{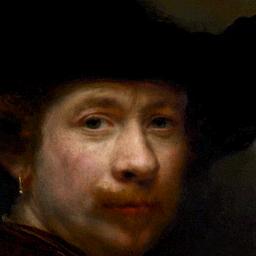}
\includegraphics[width=.09\textwidth]{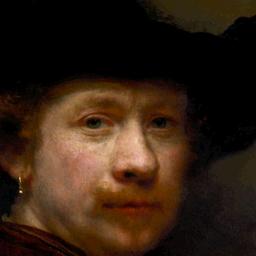}
\end{center}
  \caption{Direct interpolation between two images from the same dataset as in Figure~\ref{fig:interpolation}. Note that many intermediate images are not faces.}
\label{fig:direct_interpolation}
\end{figure*}

Image2StyleGAN \citep{Abdal2019} explored interpolations and embeddings of real images into the latent space of a GAN \citep{Goodfellow2014}. They found that while able to embed natural images almost perfectly, including those out of the distribution on which the GAN was trained, they were unable to do sensible latent space interpolations. Our method is able to do sensible interpolations between out-of-distribution datasets by interpolating the mean and covariance of their posterior distributions as we show in Figure~\ref{fig:interpolation}. This method can be thought of as first projecting the out-of-distribution images onto the flow model manifold before interpolating.

Neural Style Transfer \citep{Gatys2015} is a method for re-rendering images with a different style, while also keeping the content similar. Our method can be seen as similar to Neural Style Transfer methods, with the ``content" being provided by the flow model and the ``style" being provided by the evidence. Unlike previous Neural Style Transfer methods which work on a single content image, we learn an entire distribution from which we can sample.

CycleGAN \citep{Zhu2017} and Few-Shot Unsupervised Image-to-Image Translation \citep{Liu2019} are also methods for blending two unpaired datasets, but the aim is slightly different. Whereas these methods are capable of turning one specific image into that resembling a different class, we are able to generate many diverse samples of the class given as evidence. At the same time, our method would be unable to modify a single image in a meaningful way.

Glow \citep{Kingma2018} explored the use of manipulation vectors in order to induce specific attributes in images. Whereas their simple algebraic method required both positive and negative examples of the attribute they wished to express, we require only positive examples. We also obtain a full posterior distribution from which we can sample many diverse images, unlike their method which can only transform a single image.

Latent Constraints \citep{Engel2018} explored how to sample conditionally from an unconditional generative model, similarly to our work. The main difference is in the method employed: whereas they learn value functions in order to find regions of the latent space with both desired conditional attributes and high likelihood, we exploit the invertibility of flow models in order to find a posterior with these qualities without resorting to any gradient-based training whatsoever.

The common thread between our work and previous works is that while many previous works showed manipulations of individual images using trained generative models, we aim to combine pre-trained generative models with new data to create an entirely new generative model.

\section{Experiments}
\label{sec:experiments}

While there are many different types of flow models such as NICE \citep{Dinh2014}, Real NVP \citep{Dinh2016}, and Flow++ \citep{Ho2019}, we chose to use Glow \citep{Kingma2018} for all of our experiments. This is an arbitrary choice mainly influenced by the public availability of a pre-trained model for Glow, trained on the CelebA dataset \citep{Liu2014}. Flow models are currently prohibitively difficult to train, both in terms of time and compute requirements, and this study is solely interested in exploring transfer learning using existing models.

\subsection{In-distribution Conditioning}

\begin{figure}
\begin{center}
\includegraphics[width=.2\textwidth]{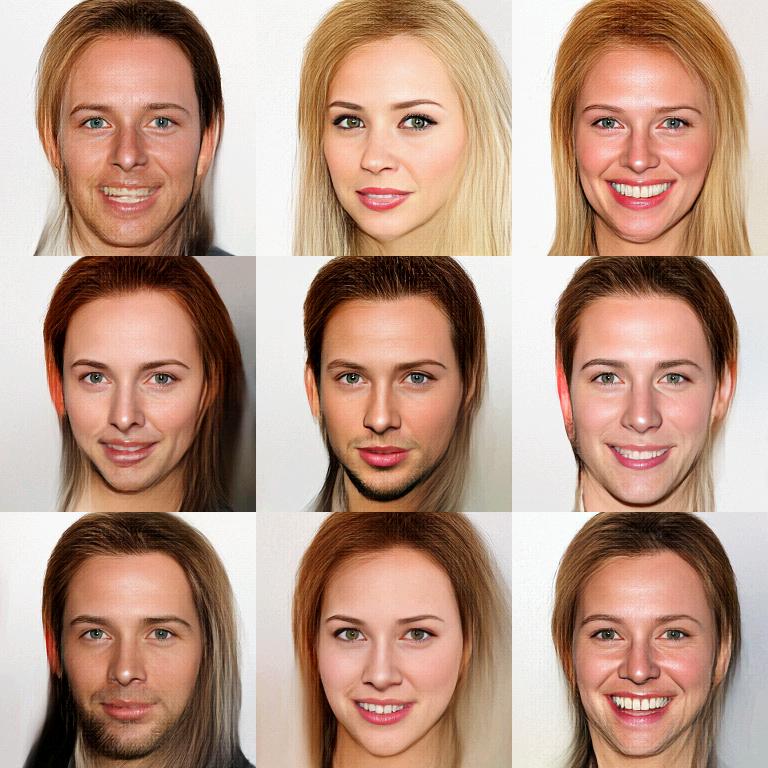}
\includegraphics[width=.2\textwidth]{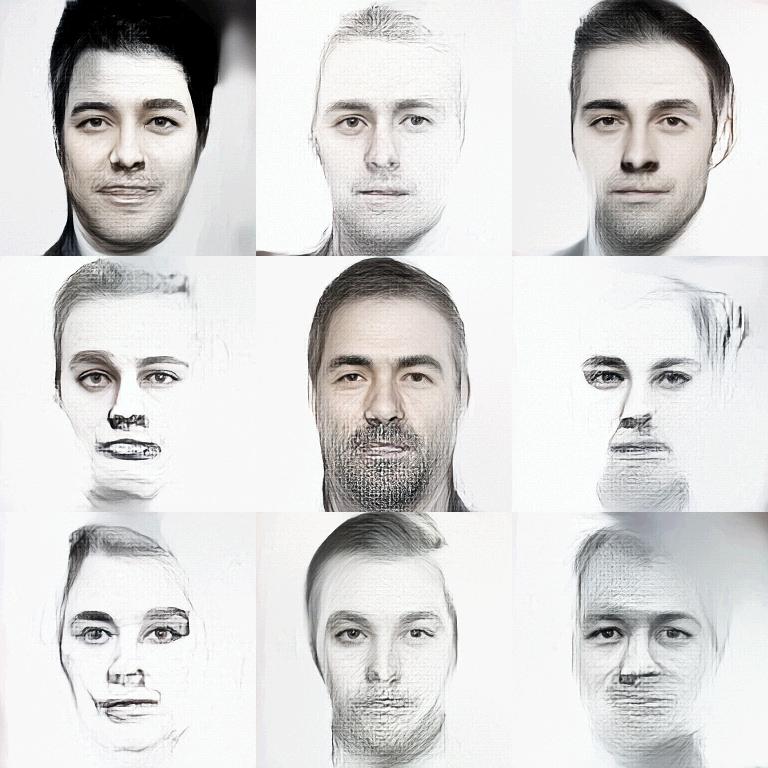}
\caption{A flow model trained on CelebA, conditioned on relatively natural images. The images of people with red hair (left) are sampled from a posterior using only five images as evidence, showing that our model is very sample-efficient for strict subsets of the CelebA distribution. Greyscale images (right), despite not appearing in the CelebA training set, were also successfully captured by a Transflow Learning posterior.}
\label{fig:in_distribution}
\end{center}
\end{figure}

A simple experiment to demonstrate the capabilities of Transflow Learning is to sample from some coherent subset of data within the CelebA dataset, such as people with red hair, people with glasses, or individual people. We found that for categories which are strict subsets of the training data, such as people with red hair, we could create a reasonable posterior distribution with both a low amount of data and a wide range of $\lambda$. In Figure~\ref{fig:in_distribution} we show results from Transflow Learning given 5 images of people with red hair, a distribution which is wholly a subset of CelebA, and 21 images of greyscale human faces, a distribution which is not represented in the CelebA training set, but is also not too far off. 

\begin{figure*}
\begin{center}
\begin{minipage}{.23\textwidth}
\includegraphics[width=\textwidth]{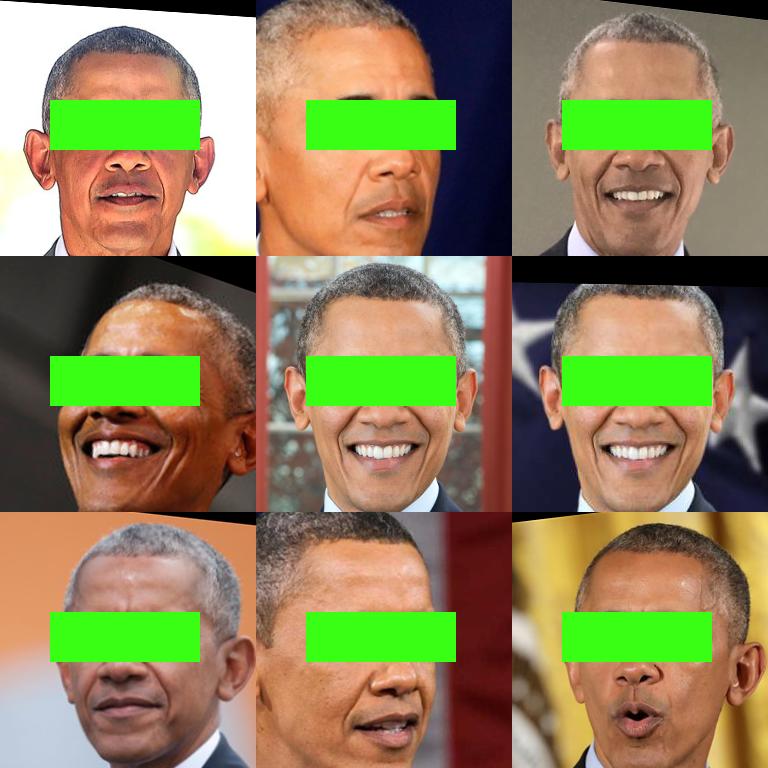}
\caption*{(a)}
\end{minipage}
\begin{minipage}{.23\textwidth}
\includegraphics[width=\textwidth]{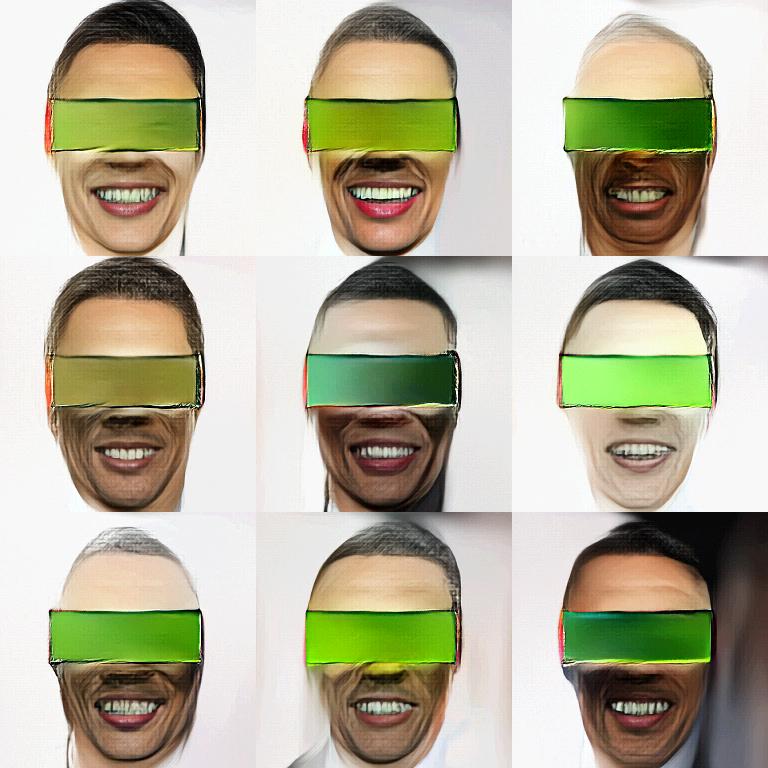}
\caption*{(b)}
\end{minipage}
\begin{minipage}{.23\textwidth}
\includegraphics[width=\textwidth]{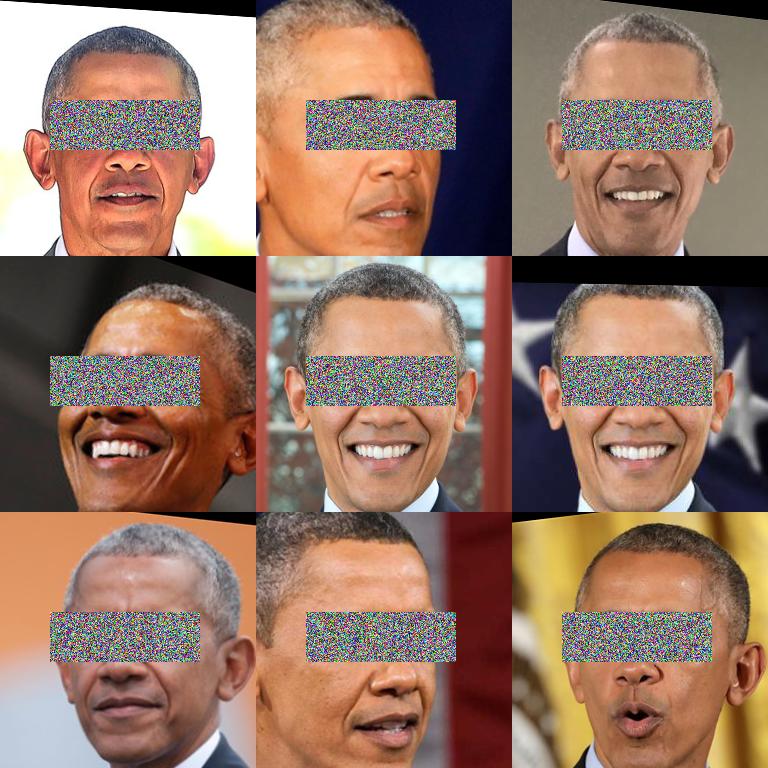}
\caption*{(c)}
\end{minipage}
\begin{minipage}{.23\textwidth}
\includegraphics[width=\textwidth]{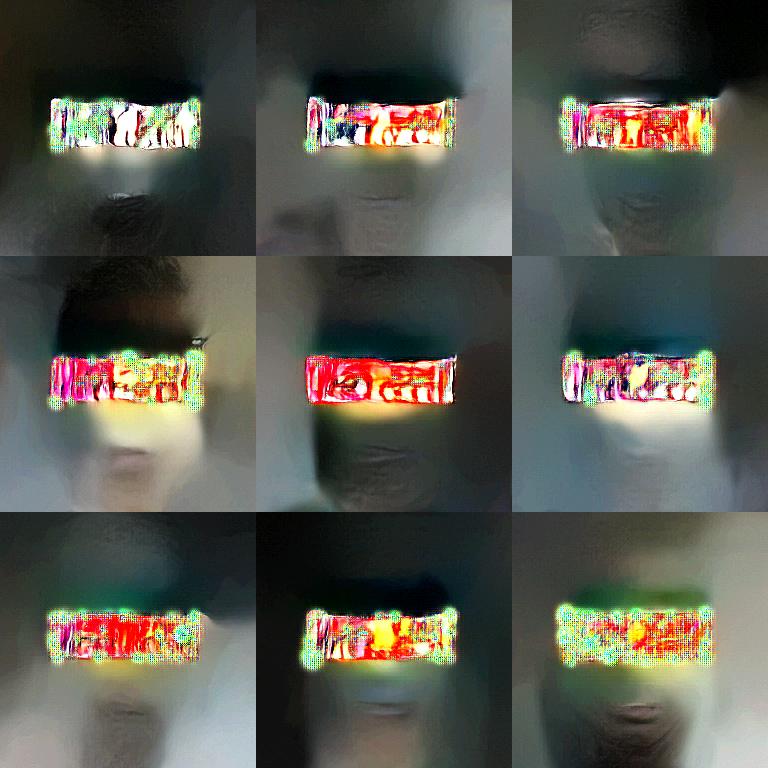}
\caption*{(d)}
\end{minipage}
\end{center}
   \caption{Even when providing Transflow learning with evidence that is far outside of the distribution on which the flow model is originally trained (a), we are able to learn a sensible posterior distribution (b). Evidence that is so unlikely that it could not have come from a natural image (c), however, causes the posterior mean to be too far from the mean of the original distribution, and output samples (d) are no longer meaningful, even for high values of $\lambda$.}

\label{fig:obama_eyes}
\end{figure*}

We also attempted to condition on natural faces with a large occlusion, and were surprised by the results. Figure~\ref{fig:obama_eyes} shows results when attemping to condition Glow on 25 images of President Obama with a large occlusion over his eyes. Transflow Learning was shockingly able to generate images of men with a neon-green occlusion over their eyes, despite similar images clearly not being located in the CelebA training set. It is important to reemphasize at this point that Transflow Learning in no way modifies the flow model---amazingly, there was simply a region in the Glow latent space in which latent vectors corresponding to these images exist, and Transflow Learning was able to find a Gaussian covering this space. As the posterior contains elements of both the prior and the evidence, we expected the posterior to perform similar to inpainting, and were surprised to learn that the latent space of Glow was rich enough to be able to generate these images which were far outside of the training set. We only observed an inpainting-like effect for relatively high values of $\lambda$ (\ie , values close to $m$), but at that point the model had forgotten to also generate President Obama, and was generating seemingly random samples with a faint, translucent occlusion around the eyes.

Even more surprisingly, when we changed the occlusion so that it would be made up of random pixels as opposed to one solid colour, Transflow Learning was no longer able to generate human faces, even for relatively high values of $\lambda$. We believe that this effect is due to how unlikely the noisy occlusion is compared to the monochromatic occlusion. We found that latent vectors corresponding to a real image of President Obama, the same image of President Obama with a monochromatic occlusion, and an image of an anime character have the log-likelihoods of -284,462, -281,377, and -285,610 respectively. Notably, these are all contained in roughly the same range, and the image of President Obama with a monochromatic occlusion was actually more likely than the image without the occlusion. Conversely, the latent vector corresponding to an image of President Obama with a noisy occlusion has a log-likelihood of -333,436, a number which is completely off the charts. This effect pushes the posterior too far out, to the point that samples around the posterior mean no longer correspond to meaningful images. Indeed, the pattern around the eyes in the posterior samples also resemble patterns that appear for any image corresponding to a latent vector with extremely high magnitude, which are guaranteed to be unlikely.

\begin{figure*}
\begin{center}
\begin{tabular}{|c|c|c|c|c|c|}
\hline
Evidence & $\lambda=0.2m$ & $\lambda=0.4m$ & $\lambda=0.6m$ & $\lambda=0.8m$ & $\lambda=m$\\
\hline
\includegraphics[width=.14\textwidth]{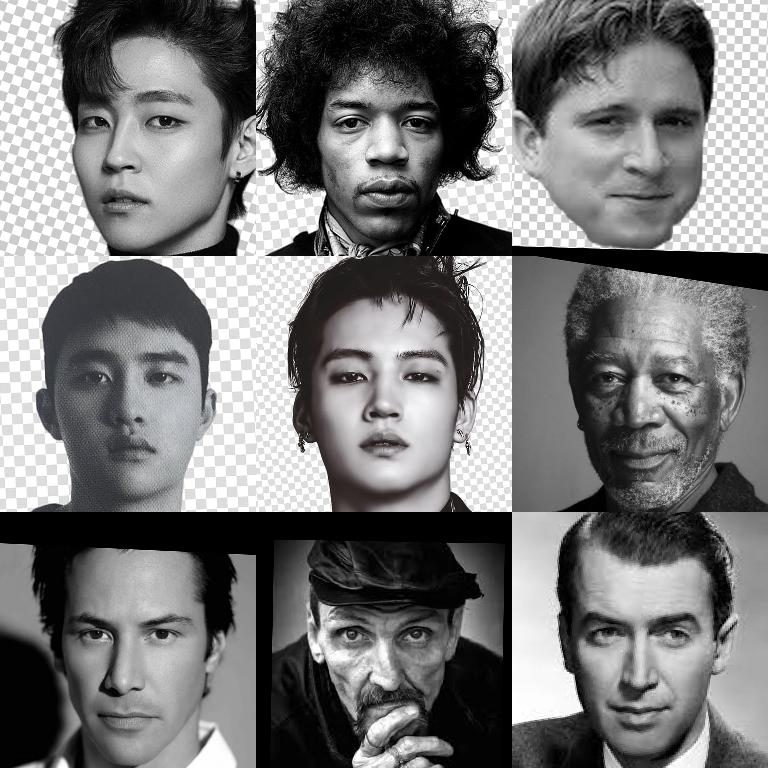} &
\includegraphics[width=.14\textwidth]{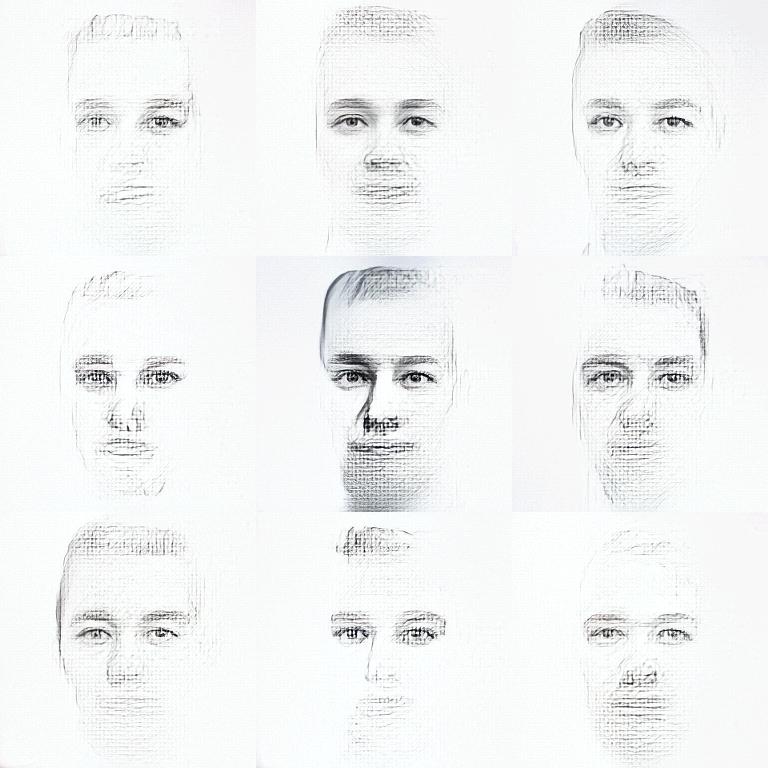} &
\includegraphics[width=.14\textwidth]{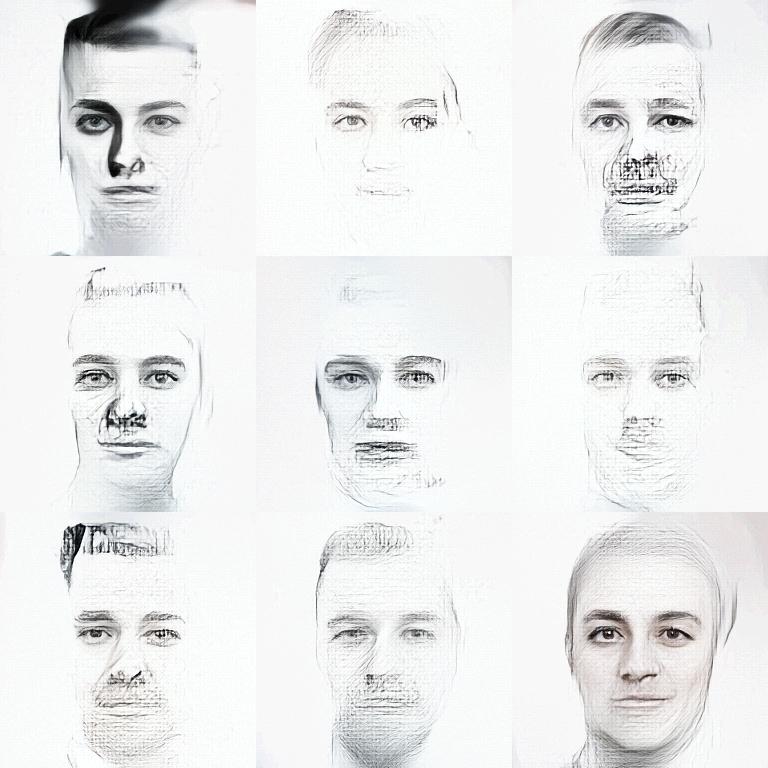} &
\includegraphics[width=.14\textwidth]{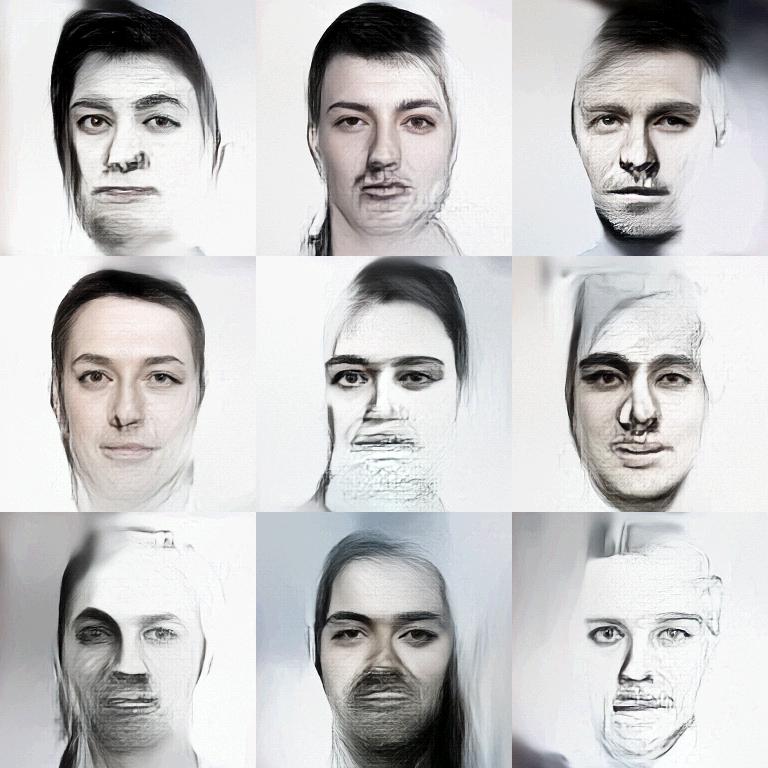} &
\includegraphics[width=.14\textwidth]{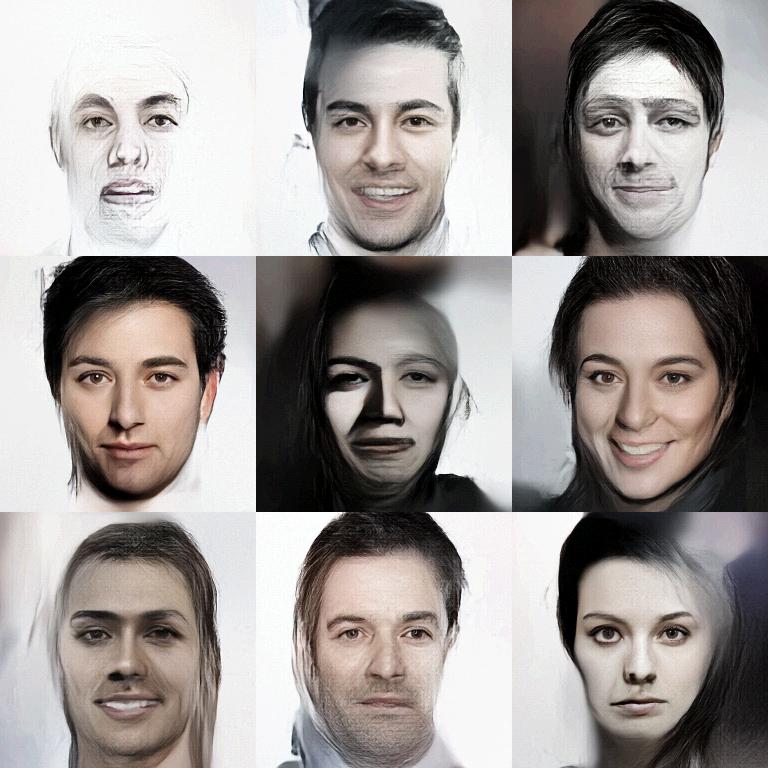} &
\includegraphics[width=.14\textwidth]{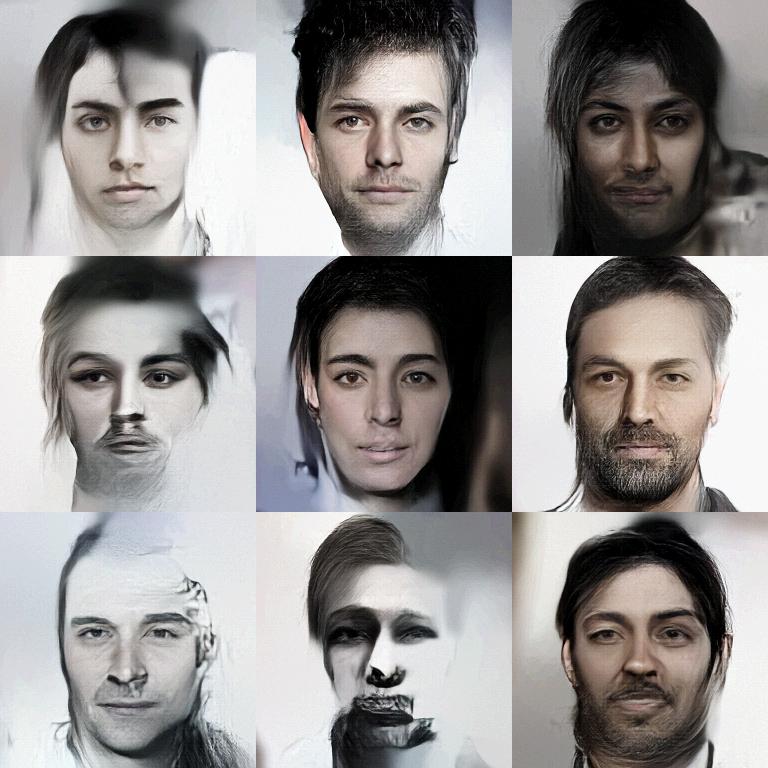}
\\
\hline
\end{tabular}
\end{center}
\caption{Varying the $\lambda$ hyperparameter for a greyscale dataset. $\lambda$ that is low creates images resembling pencil sketches, whereas $\lambda$ that is high creates images with very subdued colors.}
\label{fig:greyscale_lambda}
\end{figure*}

\subsection{Out-of-distribution Conditioning}

\begin{figure*}
\begin{center}
\begin{tabular}{|c|c|c|c|c|}
\hline
Evidence & $\lambda=0.1m$ & $\lambda=0.4m$ & $\lambda=0.7m$ & $\lambda=m$\\
\hline
\includegraphics[width=.17\textwidth]{assets/kallen_rembrandt/rembrandt_real_grid.jpg} &
\includegraphics[width=.17\textwidth]{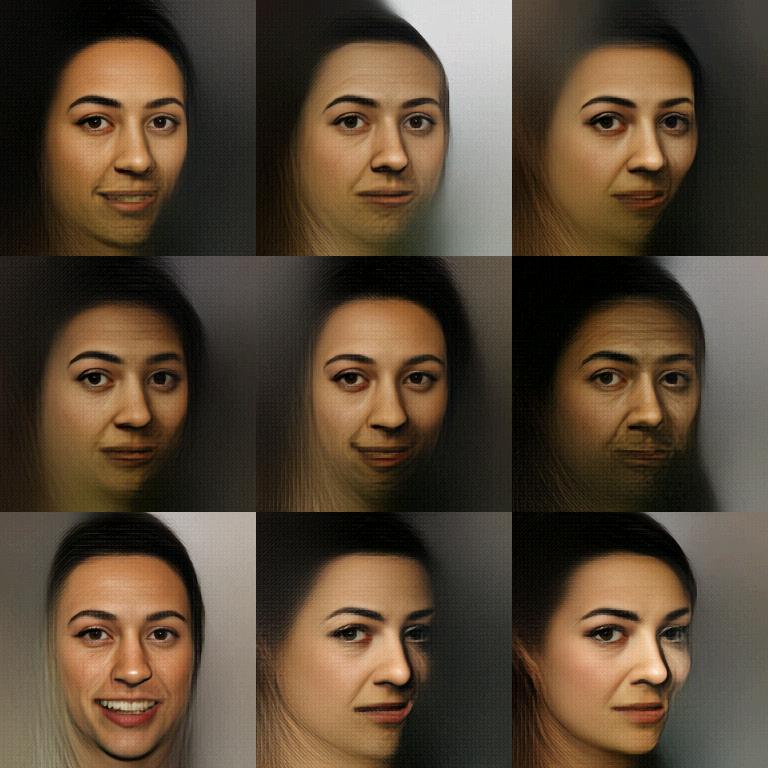} &
\includegraphics[width=.17\textwidth]{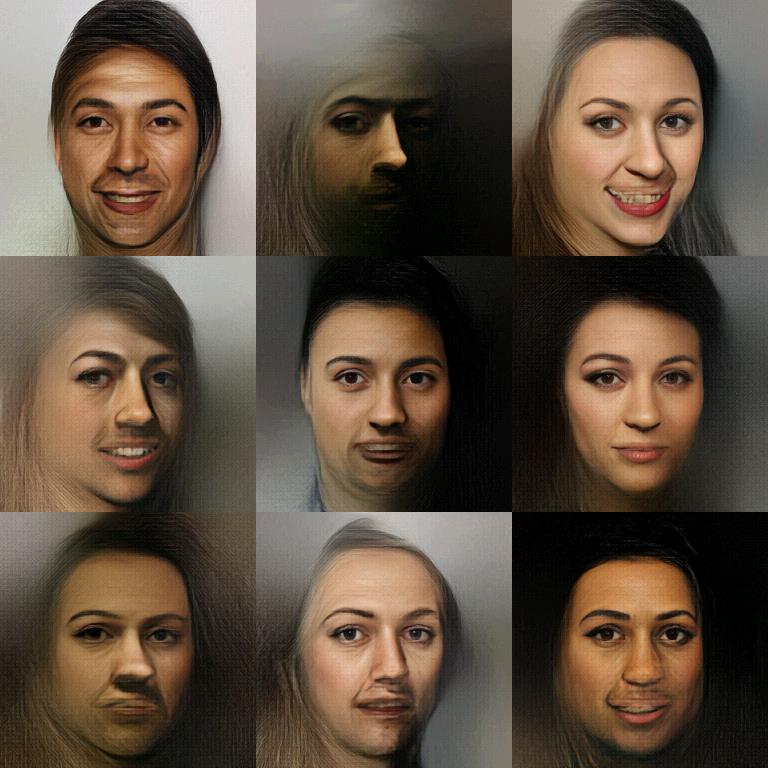} &
\includegraphics[width=.17\textwidth]{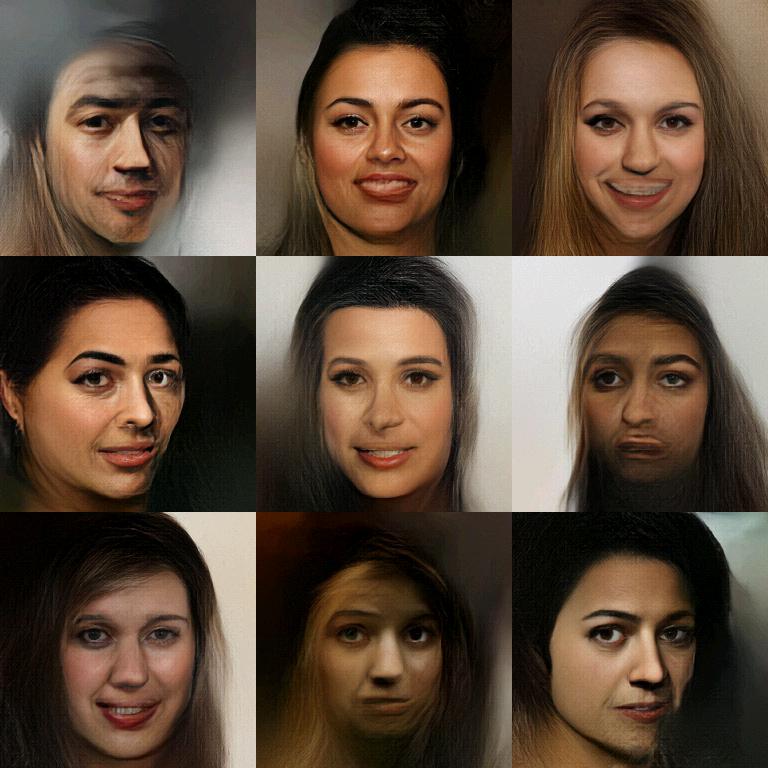} &
\includegraphics[width=.17\textwidth]{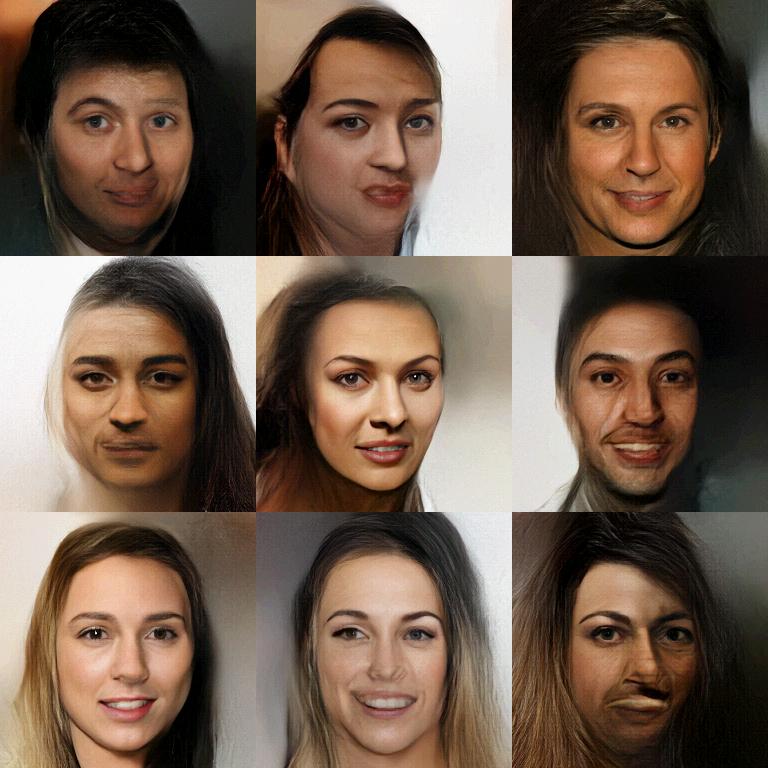}
\\
\hline
\includegraphics[width=.17\textwidth]{assets/kallen_rembrandt/kallen_real_grid.jpg} &
\includegraphics[width=.17\textwidth]{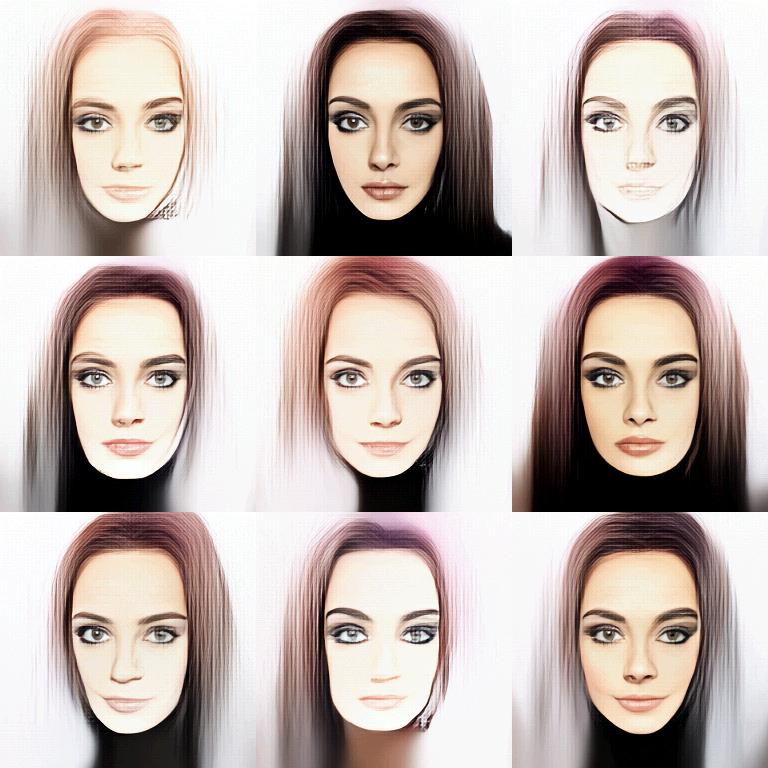} &
\includegraphics[width=.17\textwidth]{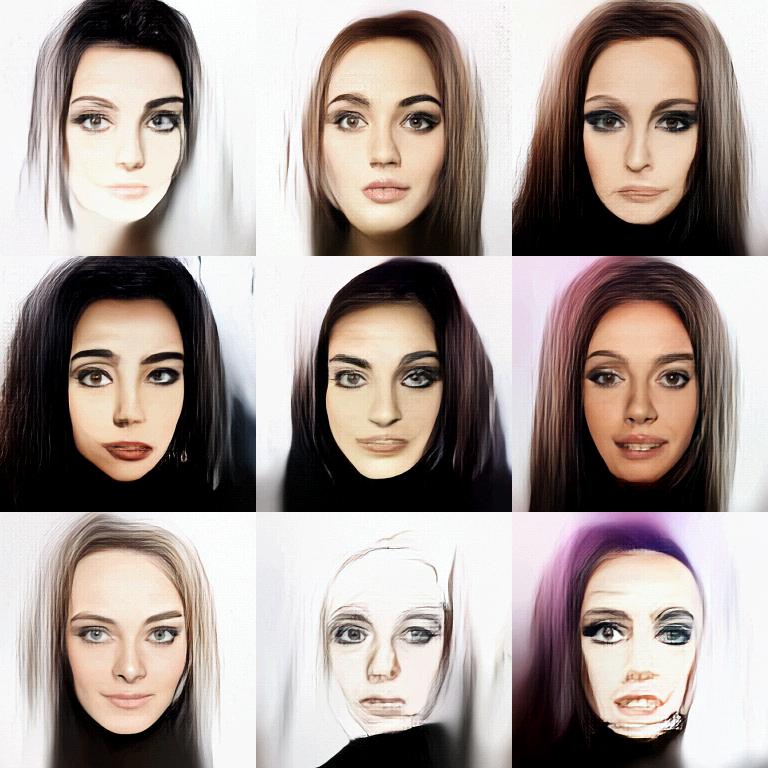} &
\includegraphics[width=.17\textwidth]{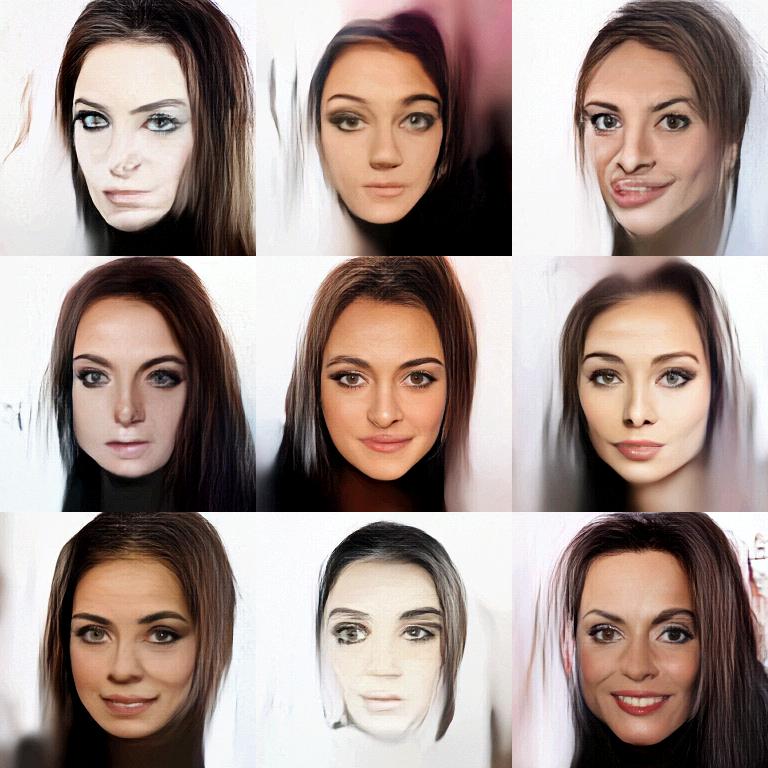} &
\includegraphics[width=.17\textwidth]{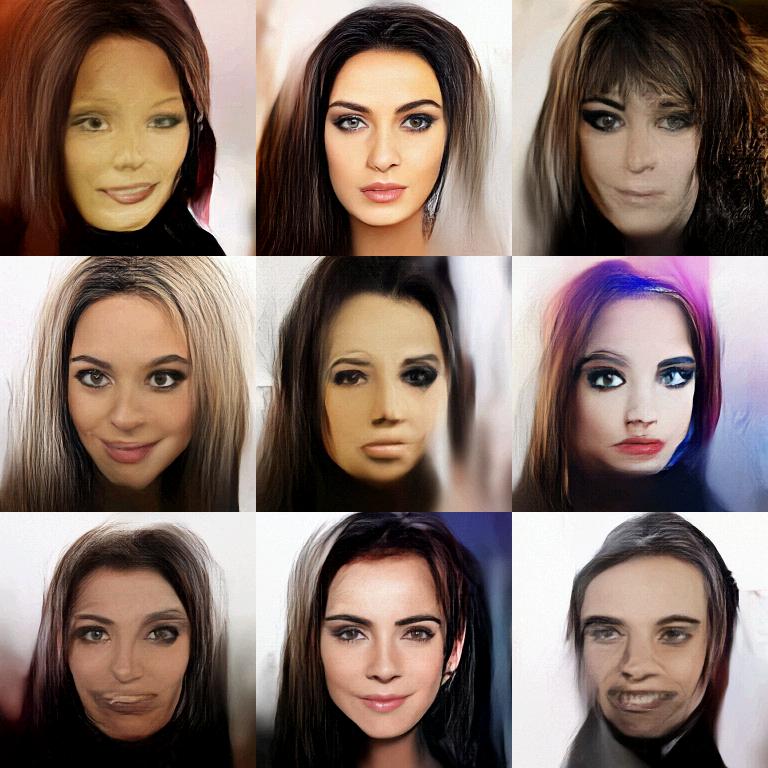}
\\
\hline
\end{tabular}
\end{center}
\caption{Varying the $\lambda$ hyperparameter for different out-of-distribution conditioning datasets. $\lambda$ that is too low creates samples too close to the sample mean, whereas $\lambda$ that is too high creates samples too close to those from the original distribution.}
\label{fig:outofdistribution_lambda}
\end{figure*}

We also found that Transflow Learning could generate samples of many types of images which are not strictly human faces. While generated images were often nonsensical when conditioned on images which could not be interpreted in any way as a face, we found that a wide variety of images, such as cartoon faces or paintings of faces, gave interesting results. In Figure~\ref{fig:outofdistribution_lambda} we show two examples of such a conditioning, on self-portraits of Rembrandt and images of an anime character.\footnote{Anime character images taken from the Anime Face Character Dataset: \url{http://www.nurs.or.jp/~nagadomi/animeface-character-dataset/}}

We found that the setting of $\lambda$ was much more difficult in out-of-distribution scenarios. While with in-distribution conditioning we could freely set $\lambda$ to any reasonable value and achieve sensible (although different) results, many settings of $\lambda$ for out-of-distribution conditioning created distributions that were either too narrow or too much like the original flow model.

The CelebA dataset \citep{Liu2014} is also strongly aligned, which created difficulty in conditioning on out-of-distribution datasets. We found that even for datasets that could be interpreted as human faces, sample quality decreased sharply in the presence of poorly aligned inputs. This posed particular difficulty when conditioning on anime faces, as the facial keypoint detector trained on human faces frequently mistook anime mouths for noses and chins for mouths, or more often failed to find a face at all.

While samples from the flow model are visually meaningless when evidence cannot be interpreted as a human face, the learned posteriors can still be used for downstream tasks. In the next section, we will show that Transflow Learning can use a flow model trained on the CelebA dataset to do MNIST classification in a low-shot setting.

\subsection{MNIST Classification}
\label{sec:mnist}

In order to classify MNIST digits through transfer learning with a pre-trained flow model, we must use the posterior predictive distribution given in Section \ref{sec:post_pred}. The workflow is as follows:

\begin{enumerate}
    \item Take a flow model pre-trained on any dataset
    \item Compute posterior predictive distributions conditioned on a number of observations from each class in MNIST, obtaining ten separate distributions
    \item When given an image of a new digit $\mathbf{x}$, compute the probability of $f^{-1}(\mathbf{x})$ under each of the ten posterior predictive distributions
    \item The new image $\mathbf{x}$ is classified as having come from the posterior predictive distribution under which it was the most likely
\end{enumerate}

We compared Transflow Learning to $k$-nearest neighbors in both pixel and the flow model latent space on the task of $m$-shot MNIST classification. Our results are located in Table \ref{table:mnist}. We wish to emphasize that unlike previous methods using generative models for few-shot learning, we did \emph{not} pre-train our flow model on the MNIST training set. For each experiment, we used the same implementation of Glow trained on CelebA and then showed each algorithm only $m$ labeled images from each class in the MNIST training set.

\begin{figure}
\begin{center}
\includegraphics[width=.15\textwidth]{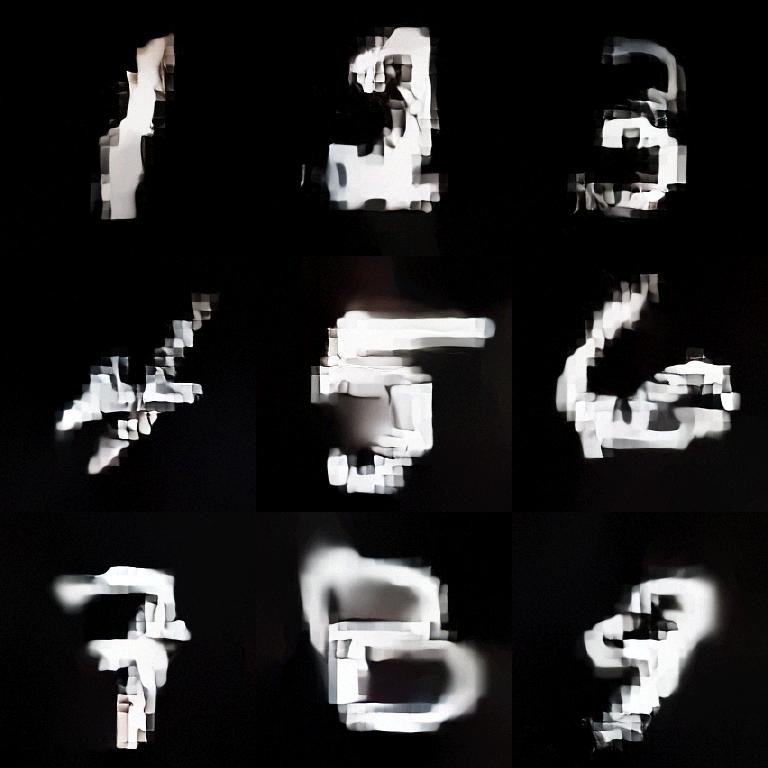}
\includegraphics[width=.15\textwidth]{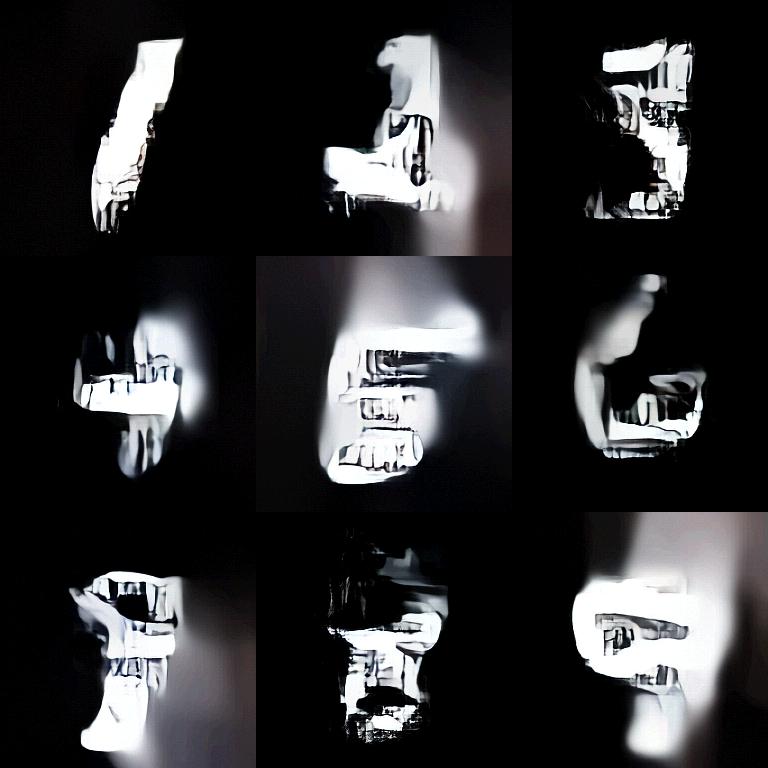}
\includegraphics[width=.15\textwidth]{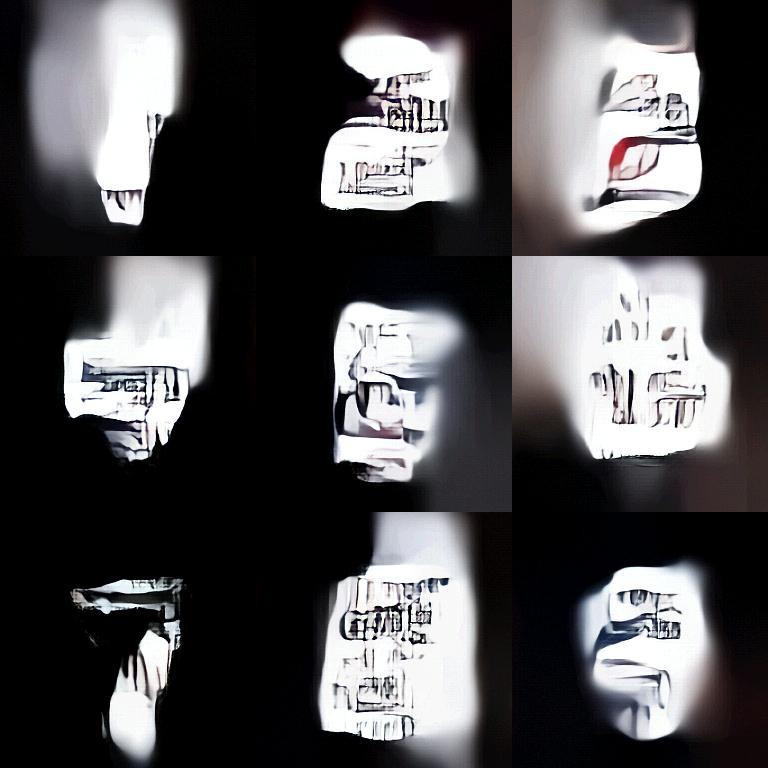}

\end{center}
\caption{Samples from the posterior of a CelebA flow model conditioned on MNIST, for $m$ equal to 1, 5, and 30 respectively. For $m$ equal to 1, the samples look very similar to the evidence. As $m$ is increased, sample quality decreases due to the sample means becoming closer to $\vec{0}$ (and therefore becoming more ``humanlike"), but prediction accuracy increases greatly.}
\label{fig:mnist_posteriors}
\end{figure}

It is also important to emphasize that no ``training" in the traditional sense is done here whatsoever. In the Transflow Learning experiments, the labeled MNIST images are simply used to warp the latent distribution of the CelebA flow model (Figure~\ref{fig:mnist_posteriors}). This has implications for transfer learning using large datasets as conditioning, as for a dataset of size $n$, we would only require $n$ evaluations of the function from data to latent variables, $f^{-1}$, in order to obtain a new classifier. Compared to the common practice of gradient-based fine-tuning of models with new training data, which requires several epochs of both costly forward and backward propagations, our method is exceptionally cheap in terms of number of function evaluations required.

\begin{table}
\begin{center}
\begin{tabular}{|c|c|c|c|}
\hline
$m$ & Ours & Pixel $k$-NN & Latent $k$-NN \\
\hline
1 & \textbf{30.39\%} & 14.99\% & 27.78\% \\
5 & \textbf{46.39\%} & 32.99\% & 35.10\% \\
10 & \textbf{58.73\%} & 19.50\% & 40.40\% \\
20 & \textbf{65.35\%} & 22.83\% & 44.12\% \\
30 & \textbf{69.52\%} & 20.55\% & 47.58\% \\
\hline
\end{tabular}
\end{center}
\caption{Accuracy on single-shot and few-shot MNIST classification, given a flow model trained on CelebA.}
\label{table:mnist}
\end{table}

\section{Future Work}
\label{sec:future_work}

While flow models are the most natural choice to study invertible generative models, it is also possible, albeit more unwieldy, to find a mapping from data to latent vectors in other generative models. One such example is the BiGAN \citep{Donahue2016}, which adds an extra term to the GAN objective in order to learn this mapping. As our methods are not specific to flow models in particular and only require the model to be invertible, it would also be possible to do posterior inference in the BiGAN latent space. As this latent space is several orders of magnitude smaller than the flow model latent space, it is very likely that posterior inference in a GAN's latent space would allow for a lower setting of the $\lambda$ hyperparameter and more finely-grained results. For instance, as the sample mean would be much closer to that of a natural image than the sample mean under a flow model, perhaps it would be possible to give multiple images of a specific person as conditioning, and generate new images of that person. At the same time, perhaps the size of the flow model latent space is contributing to the richness of samples that we are able to generate, which is a question we would like to investigate in future work.

Training generative models on multimodal datasets, such as videos with sound, is currently not feasible. If, however, it were, we could use partial data (\eg, only sound) as conditioning and then perform posterior inference in the latent space. In this scenario, the flow model would then possibly be able to generate plausible video that goes with the sound given as conditioning. Given our experiments with occluded faces, however, making this work may not be trivial.



\section{Conclusions}
\label{sec:conclusions}

We have introduced Transflow Learning, a simple method for doing transfer learning with invertible generative models. We demonstrated the capabilities of our algorithm on several generative modeling tasks, and even on downstream tasks such as handwritten digit classification.

We look forward to future research developments in invertible generative models, in particular developments in making flow models less difficult to train. Such developments would be a boon to the applicability of Transflow Learning, especially when being used for downstream tasks.


\section*{Acknowledgements}
We would like to thank Alyosha Efros, Jonathan Ho, and Jay Whang for comments on an early version of this manuscript. This work was supported by the ERC grant ERC-2012-AdG 321162-HELIOS, EPSRC grant Seebibyte EP/M013774/1 and EPSRC/MURI grant EP/N019474/1. We would also like to acknowledge the Royal Academy of Engineering and FiveAI.

{\small
\bibliographystyle{ieee_fullname}
\bibliography{cvpr}

\begin{thebibliography}{10}\itemsep=-1pt

\bibitem{Abdal2019}
Rameen Abdal, Yipeng Qin, and Peter Wonka.
\newblock {Image2StyleGAN: How to Embed Images Into the StyleGAN Latent Space?}
\newblock In {\em Proceedings of the IEEE International Conference on Computer
  Vision}, pages 4432--4441, 2019.

\bibitem{Branavan2014}
S.~R.~K. Branavan, David Silver, and Regina Barzilay.
\newblock {Learning to Win by Reading Manuals in a Monte-Carlo Framework}.
\newblock {\em Journal Of Artificial Intelligence Research}, 43:661--704, 2012.

\bibitem{Dinh2014}
Laurent Dinh, David Krueger, and Yoshua Bengio.
\newblock {NICE: Non-linear Independent Components Estimation}.
\newblock {\em International Conference on Learning Representations, Workshop
  Track Proceedings}, 2015.

\bibitem{Dinh2016}
Laurent Dinh, Jascha Sohl-Dickstein, and Samy Bengio.
\newblock {Density estimation using Real NVP}.
\newblock {\em 5th International Conference on Learning Representations, ICLR
  2017, Toulon, France, April 24-26, 2017, Conference Track Proceedings}, 2017.

\bibitem{Donahue2016}
Jeff Donahue, Philipp Kr{\"{a}}henb{\"{u}}hl, and Trevor Darrell.
\newblock {Adversarial Feature Learning}.
\newblock {\em 5th International Conference on Learning Representations, ICLR
  2017, Toulon, France, April 24-26, 2017, Conference Track Proceedings}, 2017.

\bibitem{Engel2018}
Jesse Engel, Matthew Hoffman, and Adam Roberts.
\newblock {Latent Constraints: Learning to Generate Conditionally from
  Unconditional Generative Models}.
\newblock In {\em International Conference on Learning Representations}, 2018.

\bibitem{Gatys2015}
Leon~A. Gatys, Alexander~S. Ecker, and Matthias Bethge.
\newblock {A Neural Algorithm of Artistic Style}.
\newblock {\em arXiv preprint arXiv:1508.06576}, 2015.

\bibitem{gelman2013bayesian}
Andrew Gelman, John~B Carlin, Hal~S Stern, David~B Dunson, Aki Vehtari, and
  Donald~B Rubin.
\newblock {\em {Bayesian data analysis}}.
\newblock Chapman and Hall/CRC, 2013.

\bibitem{Goodfellow2014}
Ian~J. Goodfellow, Jean Pouget-Abadie, Mehdi Mirza, Bing Xu, David
  Warde-Farley, Sherjil Ozair, Aaron Courville, and Yoshua Bengio.
\newblock {Generative Adversarial Networks}.
\newblock In {\em Advances in neural information processing systems}, pages
  2672--2680, 2014.

\bibitem{Ho2019}
Jonathan Ho, Xi Chen, Aravind Srinivas, Yan Duan, and Pieter Abbeel.
\newblock {Flow++: Improving Flow-Based Generative Models with Variational
  Dequantization and Architecture Design}.
\newblock In {\em International Conference on Machine Learning}, pages
  2722--2730, 2019.

\bibitem{Hoffman2012}
Matt Hoffman, David~M. Blei, Chong Wang, and John Paisley.
\newblock {Stochastic Variational Inference}.
\newblock {\em The Journal of Machine Learning Research}, 14(1):1303--1347,
  2013.

\bibitem{Kingma2018}
Diederik~P. Kingma and Prafulla Dhariwal.
\newblock {Glow: Generative Flow with Invertible 1x1 Convolutions}.
\newblock In {\em Advances in Neural Information Processing Systems}, pages
  10215--10224, 2018.

\bibitem{Kingma2013}
Diederik~P Kingma and Max Welling.
\newblock {Auto-Encoding Variational Bayes}.
\newblock {\em arXiv preprint arXiv:1312.6114}, 2013.

\bibitem{koch2015siamese}
Gregory Koch.
\newblock {\em {Siamese Neural Networks for One-Shot Image Recognition}}.
\newblock PhD thesis, University of Toronto, 2015.

\bibitem{lake2011one}
Brenden Lake, Ruslan Salakhutdinov, Jason Gross, and Joshua Tenenbaum.
\newblock {One shot learning of simple visual concepts}.
\newblock In {\em Proceedings of the Annual Meeting of the Cognitive Science
  Society}, volume~33, 2011.

\bibitem{LeCunYann1998}
{LeCun Yann}, {Cortes Corinna}, and {Burges Christopher}.
\newblock {THE MNIST DATABASE of handwritten digits}.
\newblock {\em http://yann.lecun.com/exdb/mnist/}, 1998.

\bibitem{Liu2019}
Ming-Yu Liu, Xun Huang, Arun Mallya, Tero Karras, Timo Aila, Jaakko Lehtinen,
  and Jan Kautz.
\newblock {Few-Shot Unsupervised Image-to-Image Translation}.
\newblock {\em arXiv preprint arXiv:1905.01723}, 2019.

\bibitem{Liu2014}
Ziwei Liu, Ping Luo, Xiaogang Wang, and Xiaoou Tang.
\newblock {Deep Learning Face Attributes in the Wild}.
\newblock In {\em Proceedings of the IEEE international conference on computer
  vision}, pages 3730--3738, 2015.

\bibitem{mansinghka2013approximate}
Vikash~K Mansinghka, Tejas~D Kulkarni, Yura~N Perov, and Josh Tenenbaum.
\newblock {Approximate bayesian image interpretation using generative
  probabilistic graphics programs}.
\newblock In {\em Advances in Neural Information Processing Systems}, pages
  1520--1528, 2013.

\bibitem{marjoram2003markov}
Paul Marjoram, John Molitor, Vincent Plagnol, and Simon Tavar{\'{e}}.
\newblock {Markov chain Monte Carlo without likelihoods}.
\newblock {\em Proceedings of the National Academy of Sciences},
  100(26):15324--15328, 2003.

\bibitem{pan2009survey}
Sinno~Jialin Pan and Qiang Yang.
\newblock {A survey on transfer learning}.
\newblock {\em IEEE Transactions on knowledge and data engineering},
  22(10):1345--1359, 2009.

\bibitem{Rezende2015}
Danilo~Jimenez Rezende and Shakir Mohamed.
\newblock {Variational Inference with Normalizing Flows}.
\newblock In {\em International Conference on Machine Learning}, pages
  1530--1538, 2015.

\bibitem{schlaifer1961applied}
Robert Schlaifer and Howard Raiffa.
\newblock {\em {Applied statistical decision theory}}.
\newblock 1961.

\bibitem{Tsividis2017}
Pedro~A Tsividis, Thomas Pouncy, Jacqueline~L Xu, Joshua~B Tenenbaum, and
  Samuel~J Gershman.
\newblock {Human Learning in Atari}.
\newblock {\em 2017 AAAI Spring Symposium Series}, 2017.

\bibitem{Oord2016}
Aaron van~den Oord, Nal Kalchbrenner, Oriol Vinyals, Lasse Espeholt, Alex
  Graves, and Koray Kavukcuoglu.
\newblock {Conditional Image Generation with PixelCNN Decoders}.
\newblock In {\em Advances in neural information processing systems}, pages
  4790--4798, 2016.

\bibitem{wilkinson2013approximate}
Richard~David Wilkinson.
\newblock {Approximate Bayesian computation (ABC) gives exact results under the
  assumption of model error}.
\newblock {\em Statistical applications in genetics and molecular biology},
  12(2):129--141.

\bibitem{Zhu2017}
Jun-Yan Zhu, Taesung Park, Phillip Isola, and Alexei~A. Efros.
\newblock {Unpaired Image-to-Image Translation using Cycle-Consistent
  Adversarial Networks}.
\newblock In {\em Proceedings of the IEEE international conference on computer
  vision}, pages 2223--2232, 2017.

\end{thebibliography}
}

\end{document}